\documentclass[superscriptaddress, reprint, amsmath, amssymb, aps, pra, floatfix]{revtex4-2}

\usepackage{mathptmx} 
\usepackage{blindtext}
\usepackage{xcolor} 
\usepackage{hyperref} 

\hypersetup{
	colorlinks=true,
	linkcolor=blue,
	filecolor=magenta,
	urlcolor=magenta,
}

\usepackage{mathrsfs}
\usepackage{graphicx} 
\usepackage{subfigure} 
\usepackage{graphicx}
\usepackage{subcaption} 
\usepackage{orcidlink}
\usepackage{comment}

\usepackage{amsmath}

\definecolor{dgreen}{rgb}{0.0,0.6,0.0}

\definecolor{pink}{rgb}{1,0,0.9}
\newcommand{\fixme}[1]{{\color{pink}#1}}

\begin{document}
\title{Physics-informed time series analysis with Kolmogorov-Arnold Networks under Ehrenfest constraints}

\author{Abhijit Sen \orcidlink{0000-0003-2783-1763}}
\email{abhijit913@gmail.com}
\affiliation{Department of Physics and Engineering Physics, Tulane University,  New Orleans, Louisiana 70118, USA}

\author{Illya V. Lukin\orcidlink{0000-0002-8133-2829}}
\email{illya.lukin11@gmail.com}
\affiliation{Karazin Kharkiv National University, Svobody Square 4, 61022 Kharkiv, Ukraine}
\affiliation{Akhiezer Institute for Theoretical Physics, NSC KIPT, Akademichna 1, 61108 Kharkiv, Ukraine}

\author{Kurt Jacobs\orcidlink{0000-0003-0828-6421}}
\email{dr.kurt.jacobs@gmail.com}
\affiliation{United States DEVCOM Army Research Laboratory, Adelphi, Maryland 20783, USA}
\affiliation{Department of Physics, University of Massachusetts at Boston, Boston, Massachusetts 02125, USA}

\author{Lev Kaplan\orcidlink{0000-0002-7256-3203}}
\email{lkaplan@tulane.edu}
\affiliation{Department of Physics and Engineering Physics, Tulane University,  New Orleans, Louisiana 70118, USA}

\author{Andrii G. Sotnikov\orcidlink{0000-0002-3632-4790}}\email{a\_sotnikov@kipt.kharkov.ua}
\affiliation{Karazin Kharkiv National University, Svobody Square 4, 61022 Kharkiv, Ukraine}
\affiliation{Akhiezer Institute for Theoretical Physics, NSC KIPT, Akademichna 1, 61108 Kharkiv, Ukraine}

\author{Denys I. Bondar\orcidlink{0000-0002-3626-4804}}\email{dbondar@tulane.edu}
\affiliation{Department of Physics and Engineering Physics, Tulane University,  New Orleans, Louisiana 70118, USA}

\begin{abstract}
The prediction of quantum dynamical responses lies at the heart of modern physics. Yet, modeling these time-dependent behaviors remains a formidable challenge because quantum systems evolve in high-dimensional Hilbert spaces, often rendering traditional numerical methods computationally prohibitive. While large language models have achieved remarkable success in sequential prediction, quantum dynamics presents a fundamentally different challenge: forecasting the entire temporal evolution of quantum systems rather than merely the next element in a sequence. Existing neural architectures such as recurrent and convolutional networks often require vast training datasets and suffer from spurious oscillations that compromise physical interpretability. In this work, we introduce a fundamentally new approach: Kolmogorov–Arnold Networks (KANs) augmented with physics-informed loss functions that enforce the Ehrenfest theorems. Our method achieves superior accuracy with significantly less training data: it requires only $5.4\%$ of the samples (200) compared to Temporal Convolution Networks (3,700).  We further introduce the Chain of KANs, a novel architecture that embeds temporal causality directly into the model design, making it particularly well-suited for time series modeling. Our results demonstrate that physics-informed KANs offer a compelling advantage over conventional black-box models, maintaining both mathematical rigor and physical consistency while dramatically reducing data requirements.


\end{abstract}

\maketitle

\section{Introduction}

The main purpose of physics is to furnish mathematical models that describe the dynamics of observables. The model should have predictive power – given data about the present and perhaps the past, it should provide an estimator of observable quantities that well approximates actual values observed in future dynamics.

While machine learning approaches can achieve initial success with time-series forecasting, they fundamentally fail to capture the underlying physical laws governing real-world systems. These black-box models often violate conservation principles, produce physically impossible predictions, and require excessive training data to rediscover known physics. Hence, the task is to develop physics-informed time-series analysis. This is easier said than done since theoretical physics deals with very different mathematical models such as, e.g., quantum and classical mechanics. It turns out that nearly all physical theories can be derived from the corresponding Ehrenfest theorems~\cite{bondar_operational_2012} -- these are dynamical constraints relating different physical observables.

For example, both non-relativistic classical as well as quantum particles moving in 1D without dissipation obey the Ehrenfest theorems~\cite{ehrenfest1927bemerkung}
\begin{align}\label{EqEhrenfestTheorems}
    \frac{d}{dt} \langle \hat{x} \rangle = \frac{1}{m} \langle \hat{p} \rangle, 
    \qquad
    \frac{d}{dt} \langle \hat{p} \rangle = \langle F(\hat{x}) \rangle,
\end{align}
where $\langle \hat{x} \rangle=\langle \hat{x} \rangle(t)$ and $\langle \hat{p} \rangle=\langle \hat{p} \rangle(t)$ are time series for a particle's position and momentum, $m$ is the particle's mass, and $F(\hat{x})$ is a shorthand notation for the sum of high-order statistical moments of the coordinate $\langle F(\hat{x}) \rangle = \sum_n c_n \langle \hat{x}^n \rangle $. Note that these differential relations cannot be solved since they do not form a closed system of equations for the unknown $\langle \hat{x} \rangle$ and $\langle \hat{p} \rangle$~\cite{ballentine1994inadequacy}. However, we can infer the classical Liouville equation of motion from the Ehrenfest theorems~\eqref{EqEhrenfestTheorems} assuming $\hat{x}$ and $\hat{p}$ commute; whereas, the Schrödinger equation is obtained if $\hat{x}$ and $\hat{p}$ do not commute~\cite{bondar_operational_2012}. This Ehrenfest-centric approach to physics has led to a number of findings in quantum control~\cite{campos_how_2017, mccaul_driven_2020}, formulating relativistic dynamics~\cite{cabrera_dirac_2016, cabrera_operational_2019}, open system theory~\cite{bondar_wignerlindblad_2016, zhdanov_no_2017}, entropic forces~\cite{vuglar_nonconservative_2018}, and entropic gravity~\cite{schimmoller_decoherence-free_2021, sung_decoherence-free_2023}. This motivates us to merge time series analysis with the Ehrenfest theorems.

Modeling how many-body quantum systems respond to external fields presents significant challenges due to their non-linear behavior, high dimensionality, and sensitivity to initial conditions. This is clearly seen in high harmonic generation (HHG), where strong input pulses create output signals at multiple frequencies of the input. HHG signals are complex because the interplay of  quantum coherence and many‑body interactions produces broadband, nonstationary waveforms with rapidly varying amplitudes and phases across multiple harmonics. Further, HHG has garnered significant interest across scientific and technological domains due to its role in generating coherent extreme ultraviolet and soft X-ray radiation. In ultrafast spectroscopy, HHG enables attosecond time resolution, thereby allowing real-time tracking of electronic dynamics~\cite{krausz_attosecond_2009}. The ability to generate high-order harmonics has catalyzed advances in nanoscale imaging and lithography~\cite{ghimire_observation_2011,luu_extreme_2015}. Recent studies indicate that HHG can enable  single-atom computing~\cite{mccaul_towards_2023} and enhance spectroscopic characterization of complex mixtures~\cite{magann_sequential_2022}. Within quantum spin systems, HHG serves as a probe for spin dynamics and holds promise for THz laser generation~\cite{takayoshi_high-harmonic_2019,ikeda_high-harmonic_2019,malla_ultrafast_2023}. Accurate modeling of such nonlinear quantum-optical processes allows for rapid prediction of system responses to arbitrary optical inputs, eliminating the need for expensive trial-and-error experimentation.

With the increasing availability of experimental and theoretical data in HHG, machine learning (ML) methods offer substantial advantages over traditional first-principles modeling. For example, Lytova et al.~\cite{Lytova2023} demonstrated that a deep neural network (DNN) can be trained to predict HHG spectra from molecular parameters (laser intensity, internuclear distance, molecular orientation) and solve the inverse problem of determining molecular parameters from observed HHG spectra or dipole data. In the macroscopic regime, Serrano et al.~\cite{serrano_machine-learning_2023} used DNN to overcome the computational impossibility of exactly simulating HHG by coupling the 3D time-dependent Schrödinger equation with Maxwell's equations, enabling macroscopic HHG simulations that reveal hidden attosecond pulse signatures missed by standard approximations. Further, ML has also been applied to HHG in broader contexts: from harmonic generation microscopy \cite{Shen2023}, nonlinear spectroscopy in solids \cite{Klimkin2023}, and plasma physics where particle-in-cell simulations were combined with ML to predict harmonic generation features \cite{Mihailescu2016}, to molecular HHG modeling and spectrum inversion \cite{Lytova2023}, time-series forecasting of nonlinear spectra \cite{Yan2022}, and prediction of HHG driven by spatially structured input fields~\cite{PablosMarn2023}. Beyond direct HHG spectral modeling, DNNs have further been used for ultrafast science tasks such as ultrashort pulse reconstruction~\cite{zahavy_deep_2018, brunner_deep_2022}, estimation of free-electron laser pulse characteristics~\cite{sanchez-gonzalez_accurate_2017}, and denoising of experimental photoelectron spectra~\cite{kumar_giri_purifying_2020}. In Ref.~\cite{sen_input-output_2025}, unlike in earlier HHG–ML work, the authors train a single ML model on many input–output signal pairs, achieving dataset-wide generalization rather than fitting a separate model for each case. The modeling involved the use of a Temporal Convolution Network (TCN), a physics-uninformed approach in the sense that the model is only made aware of the dataset and not the underlying physics principle. While effective, this approach presents key limitations (see Appendix \ref{TCN_failure}): computational overhead from convolutional matrix operations, a dependency on large datasets for reliable fitting, and the emergence of residual oscillations in the output time series, indicative of non-smooth predictions.

In this work, we fuse a state-of-the-art AI technique -- the Kolmogorov-Arnold Network~ \cite{liu2025kan} (KAN) -- with Ehrenfest theorems to develop a physically informed ML toolbox -- \emph{the KAN-Ehrenfest time-series analysis} (KAN-ETS). The versatility of this new technique is illustrated by modeling the nonlinear optical response of correlated many-body quantum systems. It is also shown that KAN-ETS outperforms the current state-of-the-art ``physics-uninformed'' time series analysis when used on the same data set.

\begin{figure}[ht]
    \centering
    \includegraphics[width=0.85\linewidth]{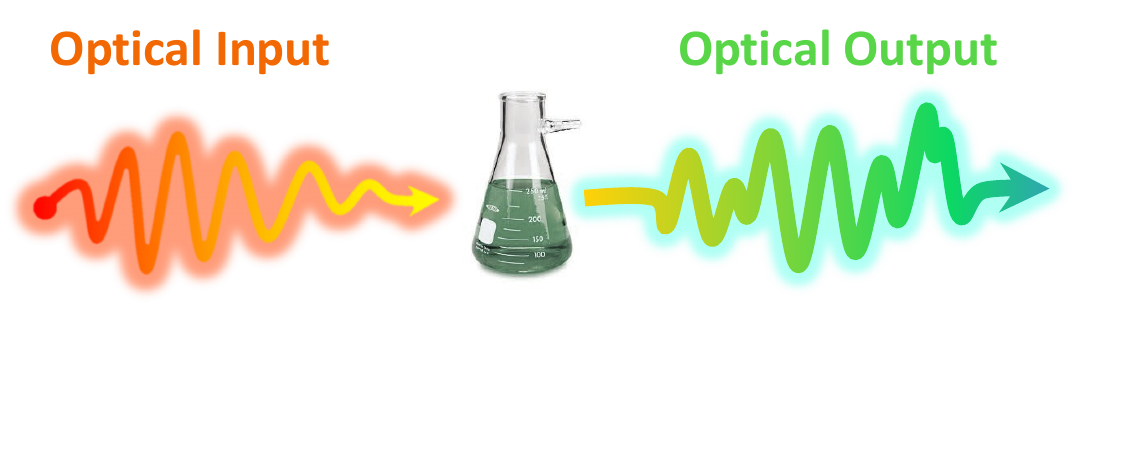}
    \includegraphics[width=0.9\linewidth]{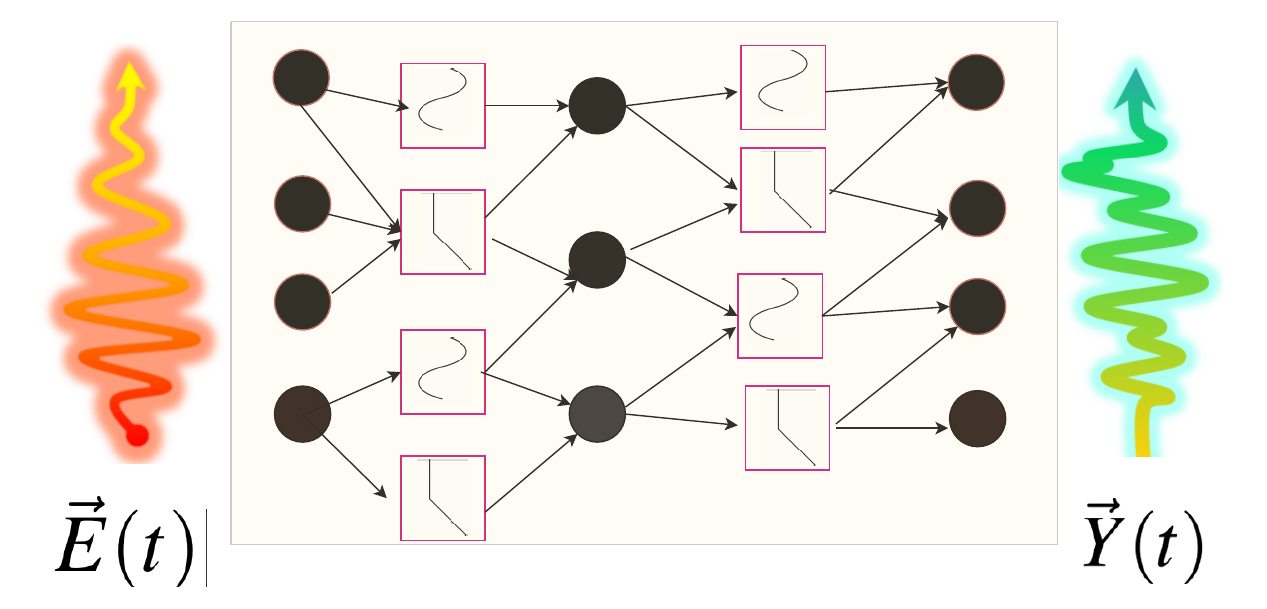}
    \caption{Optics as an input--output problem. An input field $\mathbf{E}(t)$ (red) interacts with a medium (beaker) to produce an output field $\mathbf{Y}(t)$ (green). A Kolmogorov-Arnold Network (KAN), trained on $(\mathbf{E}(t), \mathbf{Y}(t))$ pairs, serves as a surrogate model of the medium, enabling prediction of $\mathbf{Y}(t)$ for previously unseen inputs $\mathbf{E}(t)$.}
    \label{figkan}
\end{figure}


We organize the paper as follows: Section II introduces the input-output time series mapping framework and formulates the learning problem in terms of paired input and output sequences. Section III presents physics-informed Kolmogorov-Arnold Networks (KANs) with Ehrenfest constraints, establishing the theoretical foundation of our approach. Section IV describes the datasets, model architectures, loss functions, and optimization strategies that we employ in our study. Section IV also introduces the Chain of KANs architecture, which embeds temporal causality directly into the modeling process. Section V reports numerical results across single- and multi-amplitude datasets, evaluating predictive accuracy and architectural stability. Finally, Section VI concludes with a summary of our contributions and outlines potential directions for extending physics-informed KAN frameworks to broader classes of dynamical systems.

\section{Input-Output Time Series Mapping }

In input $\rightarrow$ output time series modeling, the dataset is typically organized into multiple paired sequences, each capturing a system's response to a time-varying external input or stimulus. This setup reflects a common structure in scientific problems, where understanding the temporal relationship between cause (input) and effect (output) is crucial. Each pair in the dataset corresponds to a single system run, where the input shapes the dynamics of the system and the output reflects the time evolution of relevant physical quantities.

To facilitate machine learning approaches in such continuous-time systems, both the input and output functions are discretized at uniform time steps $\delta t$. This discretization transforms the continuous input signal (say $h(t)$) into a vector $\mathbf{h} = [h_1, \dots, h_T]$ and the continuous output signal (say $Y(t)$) into a vector $\mathbf{Y} = [Y_1, \dots, Y_T]$, where $h_k = h(k\delta t)$ and $Y_k = Y(k\delta t)$. The learning task is then framed as discovering a mapping $\mathbf{h} \mapsto \mathbf{Y}$ that captures the entire dynamical transformation of the system over time. In order to train the machine learning model to do the mapping, we need an entire dataset of signals. The dataset can be expressed in the general form
\begin{equation}
\mathscr{D}(A) \; \equiv  \; \left\{\mathbf{h}^{(i)} \rightarrow \mathbf{Y}^{(i)}\right\}_{i=1}^n \,.
\end{equation}

The complete dataset $\mathscr{D}(A)$ is subsequently partitioned into training and test subsets, $\mathscr{D}_{\mathrm{train}}$ and $\mathscr{D}_{\mathrm{test}}$, to enable supervised learning and unbiased performance evaluation.

Unlike traditional autoregressive forecasting models that focus on predicting the next time step, this formulation seeks a global sequence-to-sequence transformation, allowing the model to infer complex temporal dependencies and non-linear behavior across the full input duration.


\section{Physics-Informed KAN with Ehrenfest Constraints}
Physics-Informed Neural Networks (PINNs) represent a paradigm shift in scientific machine learning by incorporating fundamental physical laws directly into the learning process of neural networks \cite{PINN1,PINNother1,PINNother2,PINNother3,PINNReview}. By embedding physical constraints through partial differential equations into the loss function, PINNs enable the learning of solutions that are both data-consistent and physically plausible. Following the revolutionary impact of PINNs and the recent introduction of Kolmogorov-Arnold Networks (KANs), the scientific machine learning community has witnessed the emergence of Physics-Informed Kolmogorov-Arnold Networks (PIKANs) \cite{PIKAN}, addressing the drawbacks of PINNs \cite{PINNremedy1,PINNdrawback,PINNdrawback1,PINNdrawback2,PINNdrawback3,PINNdrawback4,PINNdrawback5}.

KANs are grounded in the Kolmogorov–Arnold representation theorem \cite{arnold_representation_2009}, which provides exact function decomposition rather than mere approximation. Unlike DNNs that act as black boxes, KANs replace static activations with trainable spline-based edge functions, enabling compact architectures that capture the compositional structures inherent in physics. Furthermore, beyond achieving better accuracy, KANs act as a scientific discovery engine through sparsification, pruning, and symbolification; they can recover interpretable formulas from data. More details on KANs can be found in Appendix~\ref{appendix}.

Let us now delve into the general mathematical foundation of a constrained physics model for our input $\rightarrow$ output data set. Consider the Hamiltonian
\begin{equation}
H = -\frac{J_z}{4} \sum_{i=1}^{N-1} \sigma_i^z \sigma_{i+1}^z-\frac{h_x+f(t)}{2} \sum_{i=1}^N \sigma_i^x-\frac{h_z}{2} \sum_{i=1}^N \sigma_i^z \,,
\label{model_hamiltonian}
\end{equation}
where $J_{z}$ is the ferromagnetic coupling between neighboring spins, while $h_{x}$ and $h_{z}$ are the amplitudes of the transverse and longitudinal magnetic fields, respectively. The external driving $f(t)$ is considered to be our input data.

Then we have
\begin{equation}
\left\{\begin{array}{l}
H_{\text {Ising }}=-\frac{J_z}{4} \sum_{i=1}^{N-1} \sigma_i^z \sigma_{i+1}^z, \\
H_x=-\frac{h_x+f(t)}{2} \sum_{i=1}^N \sigma_i^z, \\
H_z=-\frac{h_z}{2} \sum_{i=1}^N \sigma_i^z .
\end{array}\right.
\label{Hamimltonian-components}
\end{equation}

We define the total magnetization in the $x$ direction,
\begin{equation}\label{EqDefY}
 Y_x=\sum_{i=1}^N  \sigma_i^x \,.
\end{equation}

Following Ehrenfest's theorem, the time derivative of the expectation value of any operator $A$ is given by
\begin{equation}
    \frac{d}{d t}\langle A\rangle=i\langle[H, A]\rangle+\left\langle\frac{\partial A}{\partial t}\right\rangle .
\end{equation}

In our case $A=Y_x$ does not have explicit time dependence, so
\begin{equation}
    \frac{d}{d t}\left\langle Y_x\right\rangle=i\left\langle\left[H, Y_x\right]\right\rangle \,.
    \label{ehrenfest}
\end{equation}

A quick calculation reveals the following relationship:
\begin{equation}\label{eq: ising_ehrenfest}
    \frac{d}{d t}\left\langle Y_x\right\rangle=\frac{J_z}{2} \sum_{n=1}^{N-1}\left\langle\sigma_n^y \sigma_{n+1}^z+\sigma_n^z \sigma_{n+1}^y\right\rangle+h_z  \sum_{i=1}^N \langle \sigma_i^y \rangle \,.
\end{equation}

Note that since the state $|\psi(t)\rangle$ itself evolves in time, the expectation value $
\left\langle Y_x\right\rangle (t)=\langle\psi(t)| Y_x|\psi(t)\rangle
$
is explicitly a function of $t$. It is common to write
$$
m_x(t) \equiv\left\langle Y_x\right\rangle(t)=\langle\psi(t)| \sum_{i=1}^N \sigma_i^x|\psi(t)\rangle \,.
$$

Generally, we want our neural network predictions to obey Ehrenfest equations. For this reason we will define our loss function as the sum of the minimal squared error for the signals and an additional penalty term enforcing the Ehrenfest equations. In numerical simulations, the right-hand side of the Ehrenfest equations can be computed simply using a finite difference derivative of the signal $\langle Y_{x}\rangle(t)$. But for the real experimental data, the signal is generally noisy, so it makes sense to additionally measure directly the operators appearing on the right-hand side of the Ehrenfest equations and to use them in the loss function. In this sense, the Ehrenfest equations define additional data, which does not appear in the model predictions or inputs, but helps to define the loss.




\section{Approach}

\subsection{Model and Datasets}\label{sec:Model}

To establish the correspondence between the input and output data, we study the dynamics of the one-dimensional transverse-field Ising model~\eqref{model_hamiltonian} under a time-dependent external driving field $f(t)$.


Given the external driving $f(t)$ (the input data), to obtain the output data we simulate the dynamics with the Hamiltonian~\eqref{model_hamiltonian} starting from the ferromagnetic product state in the $z$ direction. At each time step, we measure the magnetization~\eqref{EqDefY}, which corresponds to the output data. To gain access to the dynamical observables, we simulate the Schr\"odinger equation for the given Hamiltonian on infinite spin chains using the iTEBD algorithm~\cite{Vidal2007} with the maximal bond dimension $D=100$. The values of the Hamiltonian parameters are chosen to be $(J_{z}, h_{x}, h_{z}) = 0.8 \times (1, 0.25, -0.525)$.  As for the external driving, we employ the sinusoidal signals $f(t) = A \sin{(\omega t)}$ with different driving frequencies~$\omega$ and amplitudes~$A$. In all cases, we calculate the evolution for $N_{T}=500$ time steps with maximal time defined as $T = 2 \pi/\omega_{min}$, with $\omega_{min}$ the minimal driving frequency.

With these data, we should now train the KAN (or several KANs) to fit the magnetization data based on the input data~$f(t)$. Note that we employ an additional \texttt{MinMaxScaler} from \texttt{scikit-learn} on both input and output data to map it into the interval [0, 1], which is shown in several figures below. This data normalization generally improves model training and reduces possible biases from large values. 

We study the following 4 datasets: 
\begin{enumerate}
    \item\label{case1} Modeling the input $\to$ output relation for the transverse Ising model based on a precalculated dataset $\mathcal{D}(A = 2.6)$ of input-output pairs of time series
    \begin{equation}
        \mathcal{D}(A) = \{{\mathbf{h}^{(i)} \to \mathbf{Y}^{(i)}}\}, 
    \end{equation}
    where $h^{(i)}_{k} = A\sin{(\omega^{(i)}k\delta t)}$ represents the $k$th component of the $i$th sample of input vector, and where $\omega^{(i)} \in [0.4, 4]$, with either $N_\omega=200$ or $N_\omega=400$ frequencies.  
    
    \item\label{case2} Modeling the input $\to$ output relation for the transverse Ising model from dataset $\mathcal{D}(A = 10)$. Frequencies are $\omega^{(i)} \in [0.4, 4.0]$ and $N_\omega=\{200,600\}$. 
    
    \item\label{case3} Modeling the input $\to$ output relation for the transverse Ising model based on the dataset $\mathcal{D}(A_{1}=0.4, A_{m}=2.6)$ with inputs of $m=8$ different intensities,
    \begin{align}
        \mathcal{D}(A_{1}, A_{m}) & = \{\mathbf{h}^{(i, j)} \to \mathbf{Y}^{(i, j)}\}, \\
        h^{(i,j)}_{k} & = A_{j}\sin{(\omega^{(i)} k \delta t)}, \\
        A_{j} & = A_{1} + (j-1)(A_{m}-A_{1})/(m-1),
    \end{align}
    where $h^{(i,j)}_{k}$ represents the $k$th component of the $i$th sample of input vector corresponding to the amplitude $A_{j}$. Here $A_1$ and $A_m$ denote the smallest and largest amplitudes, respectively. The frequencies are taken in the range $\omega^{(i)} \in [0.4, 3]$ with $N_\omega=200$. 

    \item\label{case4} Modeling the input $\to$  output relation of the transverse Ising model on the data set $\mathcal{D}(A_{1}=1.0, A_{m}=10.0)$, $m=10$, $\omega^{(i)} \in [0.4, 4]$, and $N_\omega=400$. 
\end{enumerate}

Note that the datasets with input amplitude $A=10$ are much more complex. In addition, the trainability may depend on the number of frequencies. This is the reason for using datasets with different numbers of frequencies (while all other parameters are fixed) in some cases under study.

\subsection{Loss function and problem statement}

Our goal is to predict the output magnetization signal from a given input driving field. We discretize both input and output signals at uniform time intervals with sufficiently small time steps. The input data is represented as $\mathbf{h}^{(i)} = [h^{(i)}_{1}, h^{(i)}_{2}, \ldots, h^{(i)}_{k},\ldots,h^{(i)}_{N_{T}}]$, where $k$ indexes the time steps and $i$ indexes different input signals (corresponding to different driving frequencies and amplitudes). Similarly, we define the output signals as $\mathbf{Y}^{(i)} = [Y^{(i)}_{1}, Y^{(i)}_{2}, \ldots, Y^{(i)}_{N_{T}}]$. Our objective is to train a KAN model $\widehat{\mathbf{Y}}$ that maps input signals $\mathbf{h}^{(i)}$ to output signals $\mathbf{Y}^{(i)}$.

We construct a loss function with two components that enforce complementary constraints on the model predictions. The first term ensures that model predictions match the target output signals using a mean squared error (MSE) criterion. The second term enforces the Ehrenfest equations by requiring that the time derivatives of the model predictions satisfy the physical constraint given in Eq.~\eqref{eq: ising_ehrenfest}. In our numerical simulations, these derivatives are computed using finite differences of the simulated data, though experimentally they could be obtained through direct measurements of the operators appearing on the right-hand side of the Ehrenfest relations~\eqref{eq: ising_ehrenfest}.

The complete loss function is defined as:
\begin{align}\label{Loss}
    \mathcal{L} & = \text{MSE}(\widehat{\mathbf{Y}}(\mathbf{h}^{(i)}), \mathbf{Y}^{(i)}) \nonumber \\
     & \;\;\; + \frac{\lambda}{N_{\text{targ}} N_{T}} \sum_{i, k} |D \widehat{\mathbf{Y}}(\mathbf{h}^{(i)})_{k} - (D \mathbf{Y}^{(i)})_{k}|^{\alpha},
\end{align}
where $\lambda$ controls the weight of the Ehrenfest penalty term, $D$ denotes the finite difference approximation to the time derivative, $N_{\mathrm{targ}}$ is the number of signals in the training part of the dataset and $\alpha$ is the penalty exponent (typically 2 or 4). 

The Ehrenfest penalty term serves an additional purpose beyond physical consistency: it ensures smoothness in the temporal predictions. Without this regularization, predictions at different time points could vary independently, potentially introducing artificial noise and violating the underlying physical causality.

The hyperparameter $\lambda$ may be adjusted based on the specific system under study and can be scheduled to decrease during the optimization process to balance fitting accuracy with physical constraint satisfaction.

Both the Sobolev loss (see, e.g., Ref.~\cite{DeGaetano2025}) and Ehrenfest-constrained loss extend the standard MSE loss by incorporating derivative-based penalties, thus enforcing regularity in the learned function beyond pointwise accuracy. The Sobolev approach introduces these penalties via abstract weak derivatives, promoting smoothness in a purely mathematical sense and serving as a generic regularizer. In contrast, the Ehrenfest-constraint loss in Eq.~\eqref{Loss} replaces these abstract derivatives with physically motivated ones, specifically those derived from commutators with the system’s Hamiltonian. This substitution ensures that the enforced smoothness aligns with the underlying dynamical laws, transforming a mathematical regularizer into a physically interpretable constraint. Although both frameworks share a similar variational structure, the Ehrenfest formulation provides a principled, physics-informed specialization of the Sobolev paradigm, retaining its stability and generalization benefits while grounding every derivative penalty in fundamental physical principles.

\subsection{Model optimization}

During the model optimization step, we first divide the data set into two, using 80\% for training and the remainder for testing. Then we train the model with the \texttt{Adam} optimizer with a learning rate between 0.0002 and 0.001 with several thousand epochs depending on the problem. A brief summary of training configurations for Efficient-KANs is given in Table~\ref{tab:model}. 
\begin{table}[h!]
    \centering
\caption{Summary of training configurations.}
\label{tab:model}
    \begin{tabular}{|c|c|} \hline 
    Model type& Efficient-KAN \\ \hline
         Activation function &  SiLU \\ \hline 
         grid size &  5\\ \hline 
         spline order &  3\\ \hline 
         Optimizer &  Adam\\ \hline 
         Learning Rate &  0.0002\,$-$\,0.001\\ \hline 
         Train-Test Split &  0.8 $-$ 0.2 \\ \hline 
         Epochs &  3000$-$9000\\ \hline 
    \end{tabular}    
\end{table}

The model is then validated on the test data. In the validation process, we employ the coefficients of determination, also referred to as $R^{2}$ test scores, computed for all test cases individually. This coefficient is bounded above by 1, with $R^2\simeq1$ corresponding to a very good fit. Generally, we plot the $R^{2}$ scores as a function of the frequency of the input signal. As an additional measure of accuracy, we calculate the percentage of test cases with $R^{2}$ scores higher than some fixed threshold $R^2_{\text{th}}$, which we usually choose as 0.9, 0.95, or 0.98.

\subsection{Chain of KANs}\label{sec:ChainKAN}



In addition to employing a single KAN to predict all output points, we also test the idea of using separate KANs to calculate different data points of the output series.  Note that the main purpose of this approach is to enforce causality in the predictions, i.e., the requirement that the output at a given time step can only depend on the input at preceding time steps. Hence, we construct the estimator $\widehat{\mathbf{Y}}^{(i)} = \left[\widehat{Y}_{1}(h^{(i)}_{1}), \widehat{Y}_{2}(h^{(i)}_{1}, h^{(i)}_{2}), \widehat{Y}_{3}(h^{(i)}_{1}, h^{(i)}_{2}, h^{(i)}_{3}), \ldots, \widehat{Y}_{N_{T}}(h^{(i)}_{1}, \ldots, h^{(i)}_{N_{T}})\right]$ by minimizing the MSE loss function
\begin{align}
    \mathcal{L}_{MSE} = \frac{1}{N_{\mathrm{targ}} N_{T}}\sum_{i \in \mathrm{Train}} \, \sum_{k=1}^{N_{T}} \left[ Y^{(i)}_{k} - \widehat{Y}_{k}(h^{(i)}_{1}, h^{(i)}_{2}, \ldots, h^{(i)}_{k}) \right]^2,
\end{align}
augmented with the Ehrenfest penalty term as defined in Eq.~\eqref{Loss}. Here ``Train'' denotes the training dataset and as above $N_{\mathrm{targ}}$ is the number of signals in this dataset. Note that here any output data point prediction for $Y^{(i)}_{k}$ depends only on the input signal $h^{(i)}_{j}$ for $j \leq k$, effectively enforcing causality. Next, every function $\widehat{Y}_{k}$ can be effectively parametrized as a separate KAN. For this reasonm we refer to this structure as the ``chain of KANs.''

\section{Results} 

The results for the various models are briefly summarized in Table~\ref{tab:results2}.

\begin{table}
\caption{Performance evaluation of time series algorithms for the datasets listed in Sec.~\ref{sec:Model}, obtained from driven Ising-type systems. The
results indicate the proportion of test set predictions with $R^2>R^2_{\text{th}}$.}
\label{tab:results2}
    \begin{tabular}{|c|c|c|c|c|c|} \hline 
    Dataset & $N_\omega$ & Model& Architecture& Results& $R^2_{\text{th}}$ 
    \\ \hline
    1 & \,200\, &  Efficient-KAN & [500, 100, 500] & 39/40 & \,0.95\,
    \\ \hline
    2 & 200 &  Efficient-KAN & [500, 100, 500] & 30/40 & 0.95 
    \\ \hline 
    2 & 600 &  Efficient-KAN & [500, 400, 400, 500] & 115/120 & 0.95
    \\ \hline
    2 & 600 & Wav-KAN & [500, 240, 240, 500] & 116/120 & 0.95
    \\ \hline 
    1 & 200 & Chain of KAN & [500, 3, 1] & 37/40 & 0.95
    \\ \hline
    2 & 600 & Chain of KAN & [500, 100, 1] & 108/120 & 0.95\\ \hline
    3 & 200 & Efficient-KAN & [500, 100, 500] & 95$\%$ & 0.95\\ \hline
    3 & 200 & Chain of KAN & [500, 100, 1] & 70$\%$ & $0.9$\\ \hline
    4 & 400 & Efficient-KAN & [500, 1000, 500] & 85$\%$ & $0.9$\\ \hline
    \end{tabular}    
\end{table}

\subsection{Results for the single KAN model}

We begin our numerical analysis with the models with fully connected KAN architecture, as described in Sec.~\ref{sec:KAN}, i.e., without any causality constraints, and with datasets \ref{case1} and \ref{case2}, as specified in Sec.~\ref{sec:Model}. We discuss datasets \ref{case3} and \ref{case4} with multiple amplitudes separately in Sec.\ref{multiple_amplitudes}. 

Both our input and output data contain $500$ time steps. Hence, the corresponding KAN models generally have an architecture of the form $[500, a, 500]$, where $a$
characterizes the number of nodes in the intermediate KAN layer (we start with KAN models with a single intermediate layer and briefly describe deeper models at the end of this subsection), see also Sec.~\ref{subsec:KAN_arch}. 
Note that even one intermediate layer appears to be sufficient to approach good accuracy. Note also that the number of architecture parameters significantly depends on the number of input and output time steps. Hence, the number of parameters will be significantly reduced when treating data with lower time resolution.

\begin{figure}[h!]    
\includegraphics[width= \linewidth]{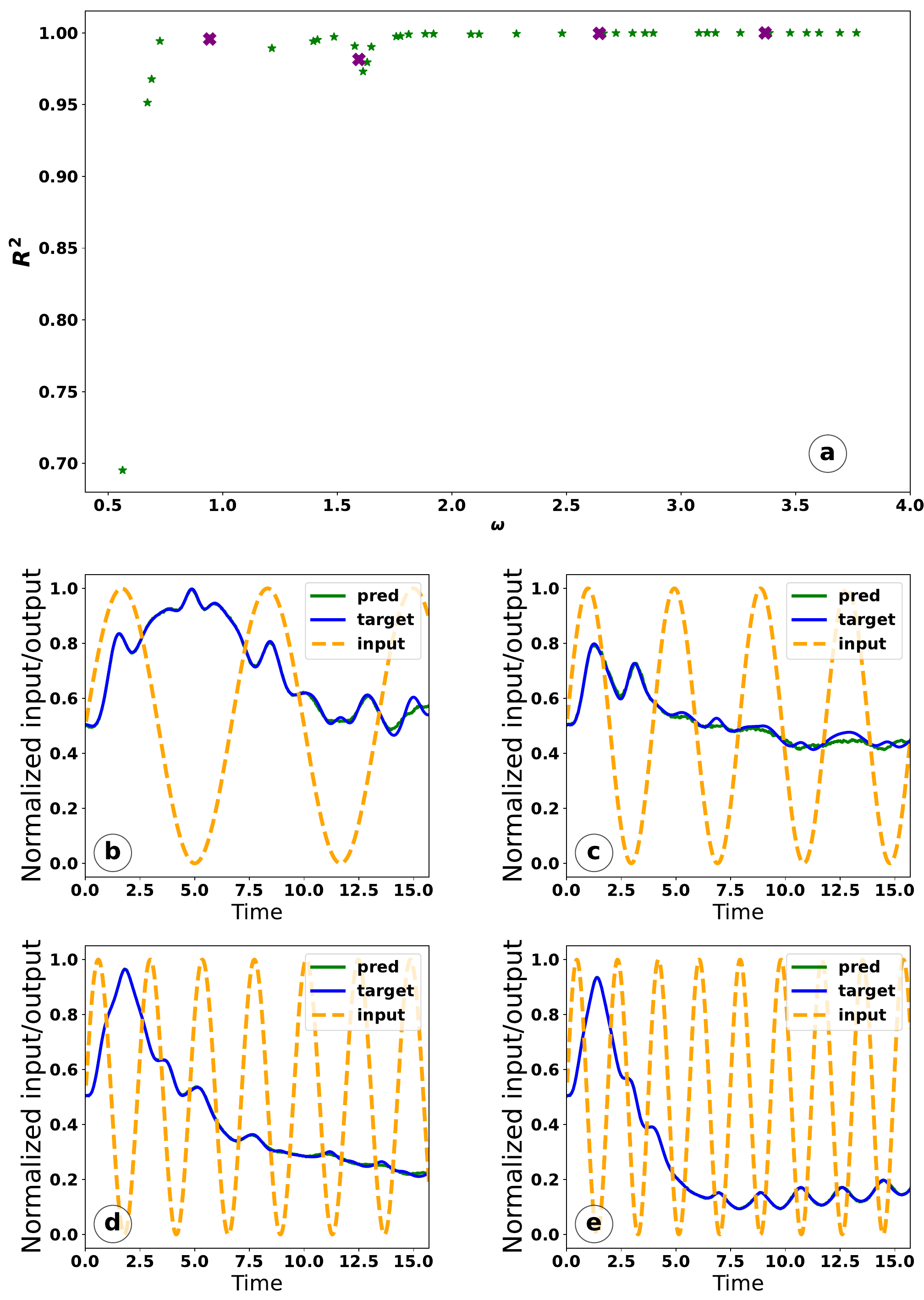}      \caption{\label{fig:SimpleKANA=2.6}%
    Performance illustration for a KAN model with intermediate layer width $100$ constructed for dataset \ref{case1} from Sec.~\ref{sec:Model} with 200 frequencies. (a) shows the $R^{2}$ values vs. the frequency $\omega^{(i)}$ for all $i$ in the test dataset. Panels (b)–(e) show the optical inputs $h_i$ (labeled as input), the corresponding output time series $Y^{(i)}$ (labeled as target), and the outputs $\hat{Y}^{(i)}$ predicted by the time series model (labeled as pred), for frequencies ${\omega^{(i)}}$ in $\{0.94, 1.59, 2.64, 3.37\}$, respectively, marked by purple crosses in panel (a).}     
\end{figure}

We first train the model on dataset \ref{case1} from Sec.~\ref{sec:Model}, which corresponds to the amplitude $A=2.6$ and $N_\omega=200$. This dataset is the easiest to learn, since the signals are generally not very complex. In Fig.~\ref{fig:SimpleKANA=2.6} we show the $R^{2}$ scores for different test samples and also display the target signal along with the KAN predictions for 4 different input frequencies. 
We can see that the predictions are generally rather good, with the only exception being at the lowest frequency, where the prediction $R^{2}$ score is near $0.7$. For all other cases, the scores are greater than $0.95$. At high frequencies, we generally obtain very precise fits, which is partially related to the fact that the output signal behavior at high frequencies is more consistent from one frequency to another. Note that here the results are presented for the KAN model with rather high internal width $a=100$. For higher widths and depths of KAN, the accuracy of the data does not change significantly. We discuss the accuracy of KAN models with lower widths (in some sense, the minimal KAN architecture) in Sec.~\ref{ArchitectureStability}. 

\begin{figure}[t]    
\includegraphics[width= \linewidth]{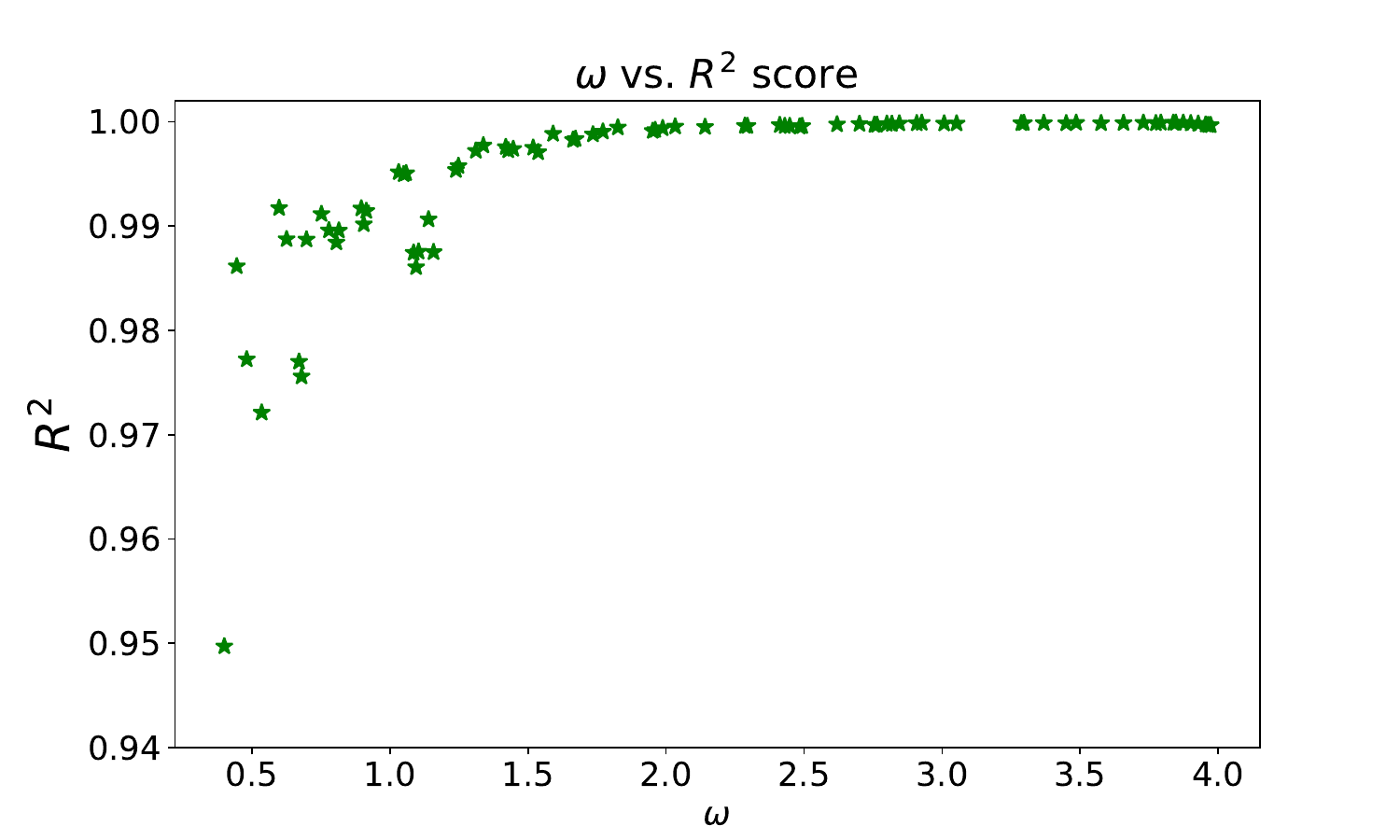}      \caption{\label{fig:Larger_dataset}%
       Performance illustration for a KAN model with intermediate layer width $100$ for dataset \ref{case1} from Sec.~\ref{sec:Model} with 400 frequencies. The figure shows the $R^{2}$ score for each driving frequency $\omega^{(i)}$ in the test set.}     
\end{figure}

Note that the prediction accuracy at the lowest frequencies can be enhanced by 
giving the model more input time series, thereby covering more input frequencies in the same range and providing higher frequency resolution. In particular, in Fig.~\ref{fig:Larger_dataset} we show the results of the KAN model with the same width $a=100$, but with a larger dataset consisting of $N_\omega=400$ frequencies and the same amplitude $A=2.6$.
Here we see that the accuracy of the predictions at low frequencies is significantly improved, as is generally expected in machine learning when the size of the training dataset increases. In particular, we find that the $R^{2}$ scores exceed $0.95$ at all frequencies in the dataset.  Note that in the previous study~\cite{sen_input-output_2025}, a dataset size $N_\omega=3700$ was necessary to accurately train the models. In contrast, for KAN models, much smaller dataset sizes ($N_\omega=200$ to $400$) are sufficient to obtain accurate results. 

\begin{figure}[h!]    
\includegraphics[width= \linewidth]{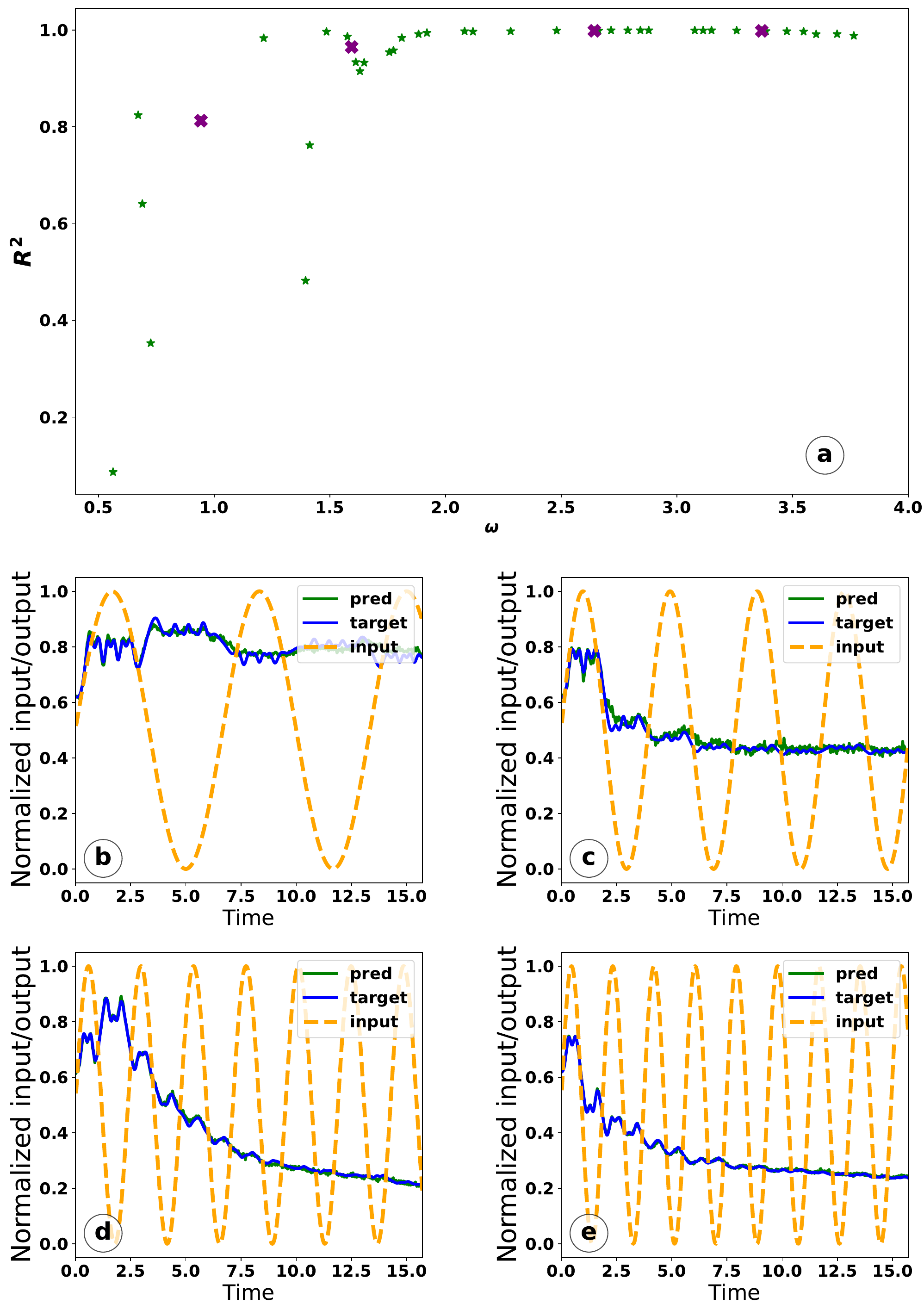}      \caption{\label{fig:SimpleKANA=10}%
       Performance illustration of a KAN model with intermediate layer width $100$ trained on  dataset \ref{case2} from Sec.~\ref{sec:Model} with 200 frequencies.  See the caption of Fig.~\ref{fig:SimpleKANA=2.6} for further details. }     
\end{figure}

Next, we study dataset \ref{case2} from Sec.~\ref{sec:Model}, which corresponds to driving with the amplitude $A=10$. The signals from this dataset are generally more difficult to learn, and larger architectures are necessary. The results for this dataset (and with $N_\omega=200$) are shown in Fig.~\ref{fig:SimpleKANA=10}. The KAN model is the same as previously used for dataset \ref{case1} with intermediate layer width $a=100$. We see that the quality of the results is poorer than in the case of lower amplitude $A=2.6$. As before, the predictions at higher frequencies are sufficiently accurate, while some discrepancies arise at small frequencies. Due to these errors at small frequencies, only 33 of 40 test cases have $R^{2}$ scores higher than 0.9, and 30 test cases have $R^{2}>0.95$. The quality of the results can be further improved by a finer frequency resolution and a larger KAN architecture. From the comparison of these two datasets, we conclude that the complexity of the data can significantly alter the necessary KAN architecture and frequency resolution for an accurate fit. 

\begin{figure}[th!]    
\includegraphics[width= \linewidth]{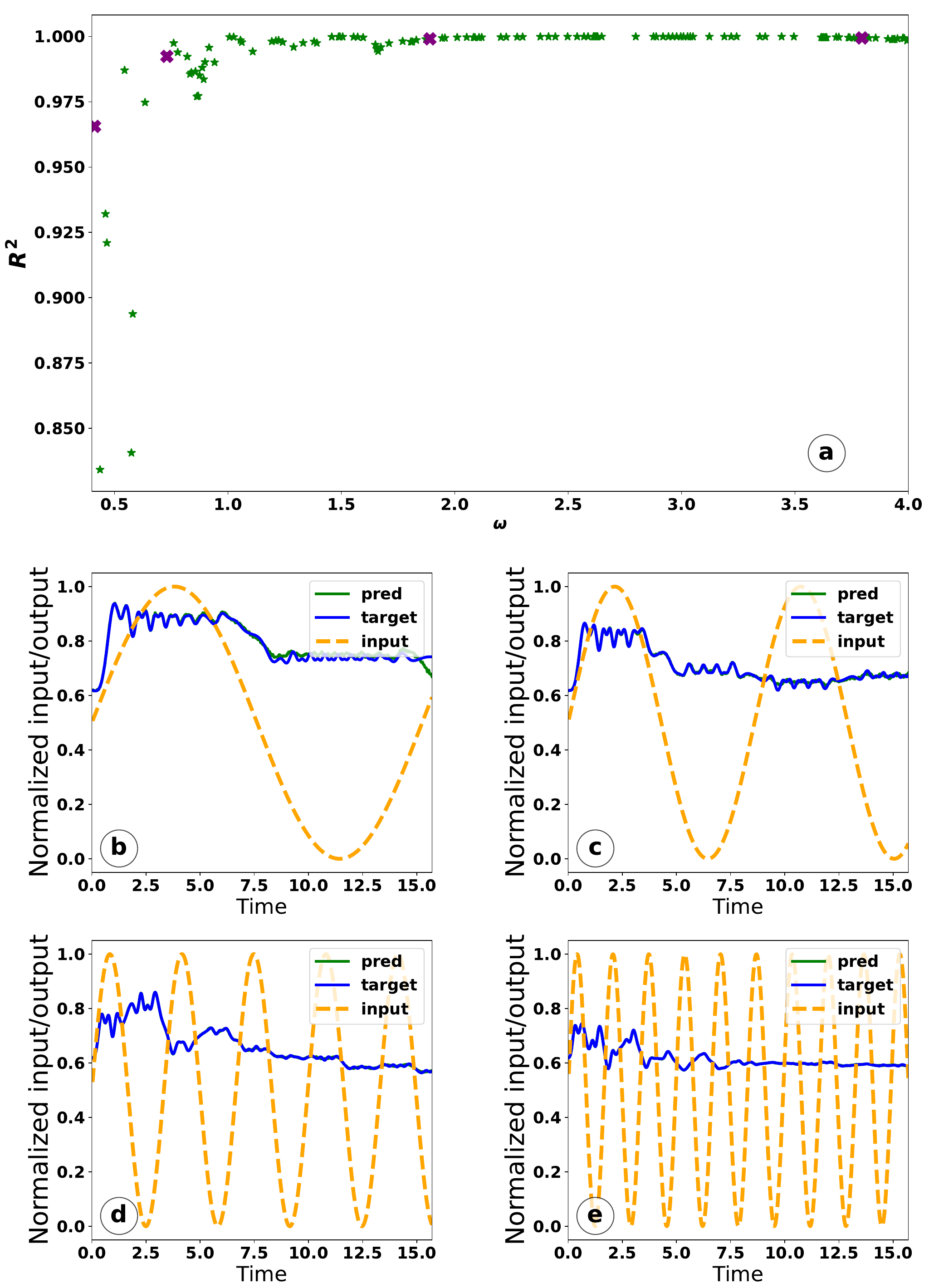}      \caption{\label{fig:deep_KANA=10}%
       Performance illustration for a 2-layer KAN with intermediate widths $400$ trained on dataset \ref{case2} from Sec.~\ref{sec:Model} with 600 frequencies. For further details, see the caption of Fig.~\ref{fig:SimpleKANA=2.6}. As the test set for the 600 frequency dataset was somewhat different, in (b-e) we show frequencies $\omega^{(i)} \in \{0.41, 0.73, 1.89, 3.80\}$.  }     
\end{figure}

\begin{figure}[h!]    
\includegraphics[width= \linewidth]{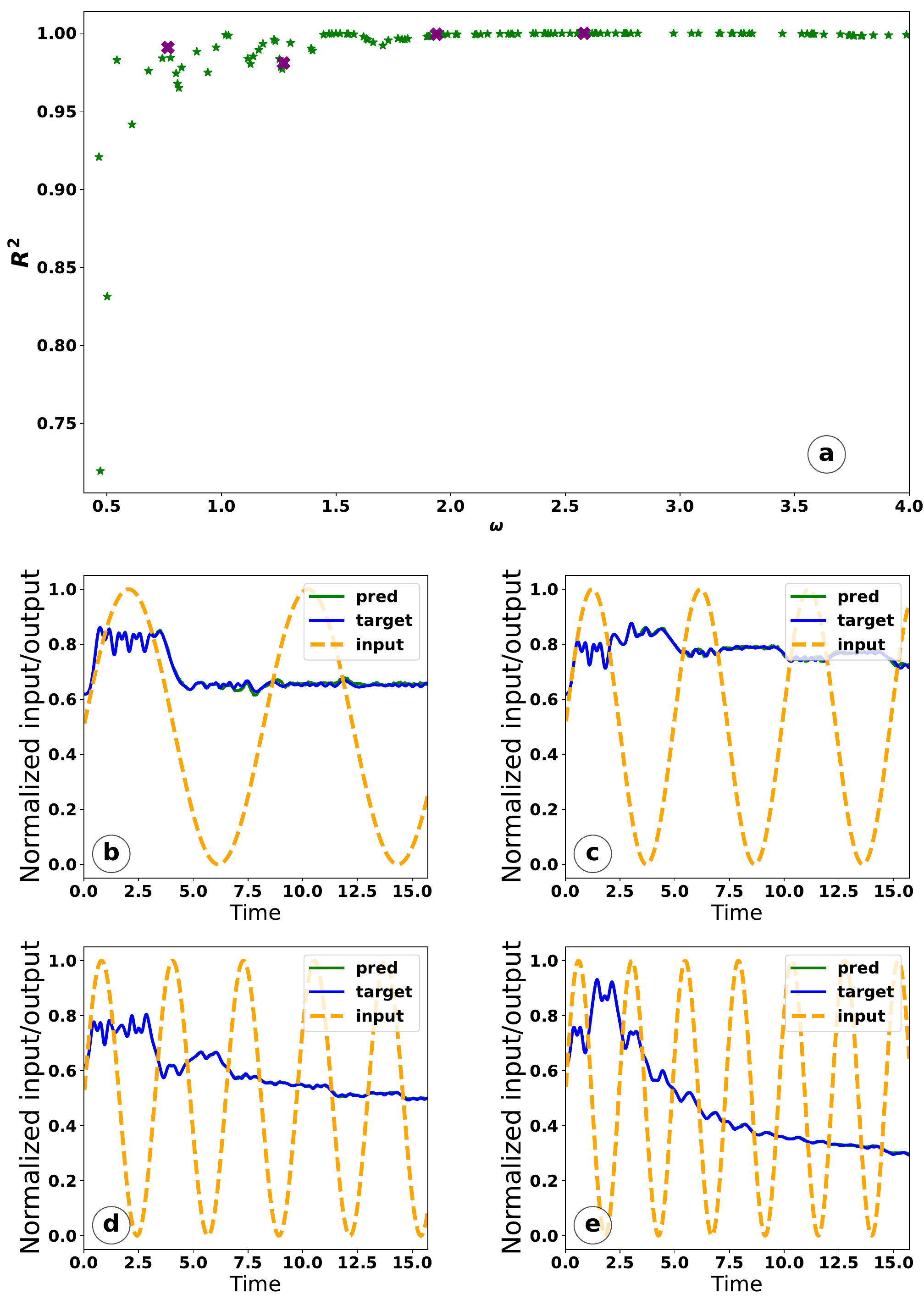}      \caption{\label{fig:waveKANA=10}%
       Performance illustration for Wav-KAN with layers $[500, 240, 240, 500]$ trained on dataset \ref{case2} from Sec.~\ref{sec:Model} with 600 frequencies. For further details, see the caption to Fig.~\ref{fig:SimpleKANA=2.6}. The frequencies shown in (b-e) correspond to $\omega^{(i)} = \{0.77, 1.27, 1.94, 2.58\}$. }     
\end{figure}

Finally, to validate the potential ability of KANs to capture these complex correlations, we train a deep KAN network with widths $[500, 400, 400, 500]$ on dataset \ref{case2} with $N_\omega=600$. The results can be seen in Fig.~\ref{fig:deep_KANA=10}. Here, 115 out of 120 test cases have $R^{2}>0.95$ and 117 cases have $R^{2}>0.9$. Note that the accuracy is much better than in the previous cases, but this is achieved at the price of larger width, depth, and dataset size. We have also tested the Wav-KAN architecture with layers $[500, 240, 240, 500]$ on the same problem and obtained the results shown in Fig.~\ref{fig:waveKANA=10}. Here, 116 out of 120 test cases have $R^{2}>0.95$ and 118 test cases have $R^{2}>0.9$. Generally, both normal spline-based Efficient-KAN and Wav-KAN are capable of achieving very high accuracy given enough training data and sufficient width. 

\subsection{Results for the chain of KANs model}

\begin{figure}[h!] 
\includegraphics[width= \linewidth]{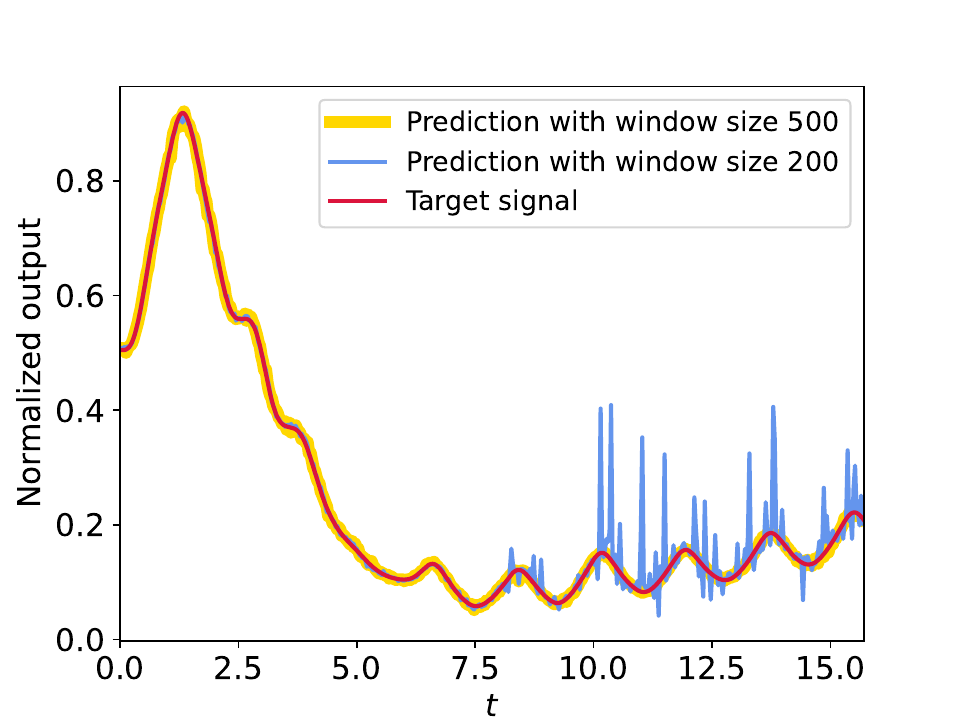}      \caption{\label{fig:ChainKANWindow}%
       Prediction and target signal for a chain of KANs model with intermediate layer width $3$ for frequency $\omega=3.55$ from dataset \ref{case1} from Sec.~\ref{sec:Model}. Fits are shown for different window sizes.  }     
\end{figure}

Next, we can discuss the chain of KANs model, which consists of a different KAN for every output point. This model was introduced in Sec.~\ref{sec:ChainKAN}. An example of predictions made using a chain of KANs with an intermediate layer of width 3 trained on dataset~\ref{case1} is shown in Fig.~\ref{fig:ChainKANWindow}. Note that here we plot the results for a chain of KANs both with full window size (500) and with restricted window size (200). It can be seen that some residual unphysical peaks may appear in the prediction, especially at small window sizes. Due to this effect, below we generally use the full window size of $500$.

\begin{figure}[h!]  
\includegraphics[width= \linewidth]{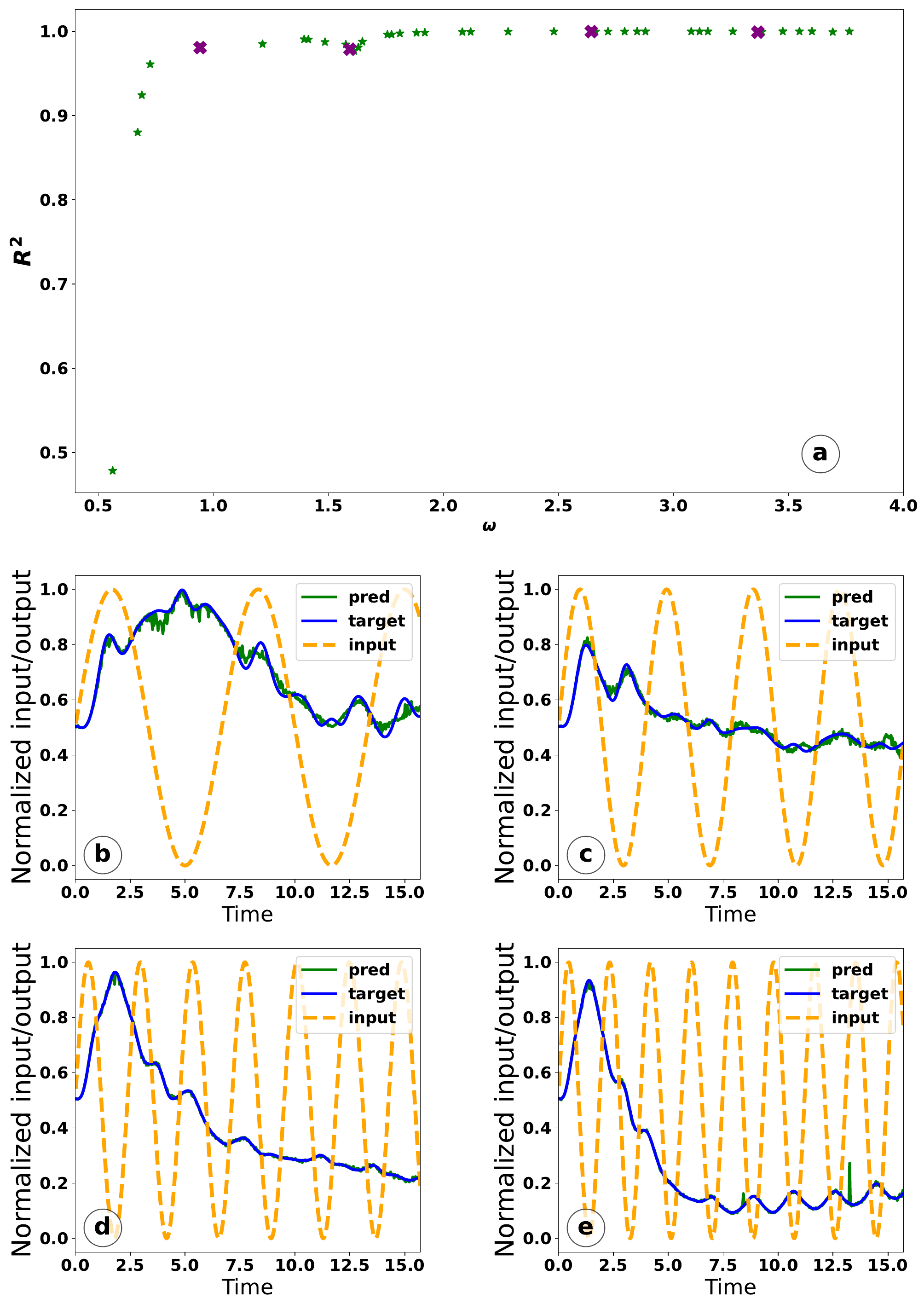}      \caption{\label{fig:ChainKANFrequency}%
    Performance illustration for the chain of KANs model with intermediate layer width $3$ trained on dataset \ref{case1} from Sec.~\ref{sec:Model}. For further details, see the caption of Fig.~\ref{fig:SimpleKANA=2.6}.}     
\end{figure}

We now discuss the prediction accuracy of the chain of KANs for datasets \ref{case1} and \ref{case2} from Sec.~\ref{sec:Model}. The results for  dataset \ref{case1} (amplitude $A=2.6$) are shown in Fig.~\ref{fig:ChainKANFrequency}. Of 40 test cases, 37 have scores $R^{2}>0.95$ and 38 cases have $R^{2}>0.9$. The results are generally similar to the single KAN model from the previous subsection, with the only inaccurate point corresponding to the lowest frequency. 

\begin{figure}[t]    
\includegraphics[width= \linewidth]{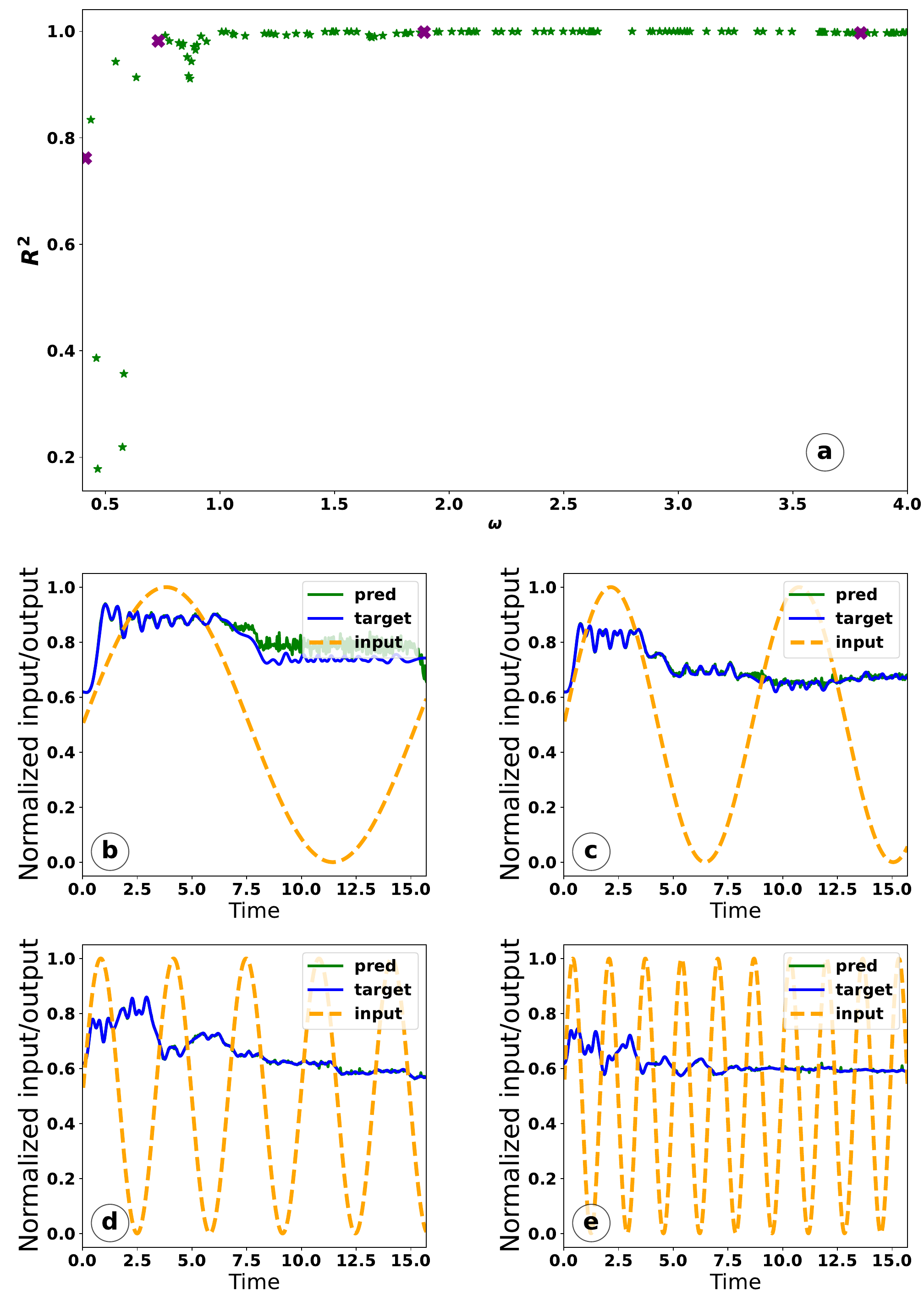}      \caption{\label{fig:ChainKANA´10}%
       Performance illustration for the chain of KANs model with intermediate layer width $100$ trained on dataset \ref{case2} with 600 frequencies from Sec.~\ref{sec:Model}. See the caption of Fig.~\ref{fig:SimpleKANA=2.6} for further details. Note that frequencies in (b-e) are the same as in Fig.~\ref{fig:deep_KANA=10}.  }     
\end{figure}

For dataset \ref{case2}, which corresponds to large-amplitude driving ($A=10$), the results are shown in Fig.~\ref{fig:ChainKANA´10}. Unfortunately, it was difficult to train the chain of KANs model for 200 frequencies and we were able to train it successfully only for a large dataset of $600$ frequencies. 108 out of 120 test cases have $R^{2}>0.95$ and 114 cases have $R^{2}>0.9$. In addition, the necessary width of the intermediate layer was also much larger than for dataset \ref{case1}.  

\subsection{Results for the datasets with multiple amplitudes} \label{multiple_amplitudes}

\begin{figure}[h!]    
\includegraphics[width= \linewidth]{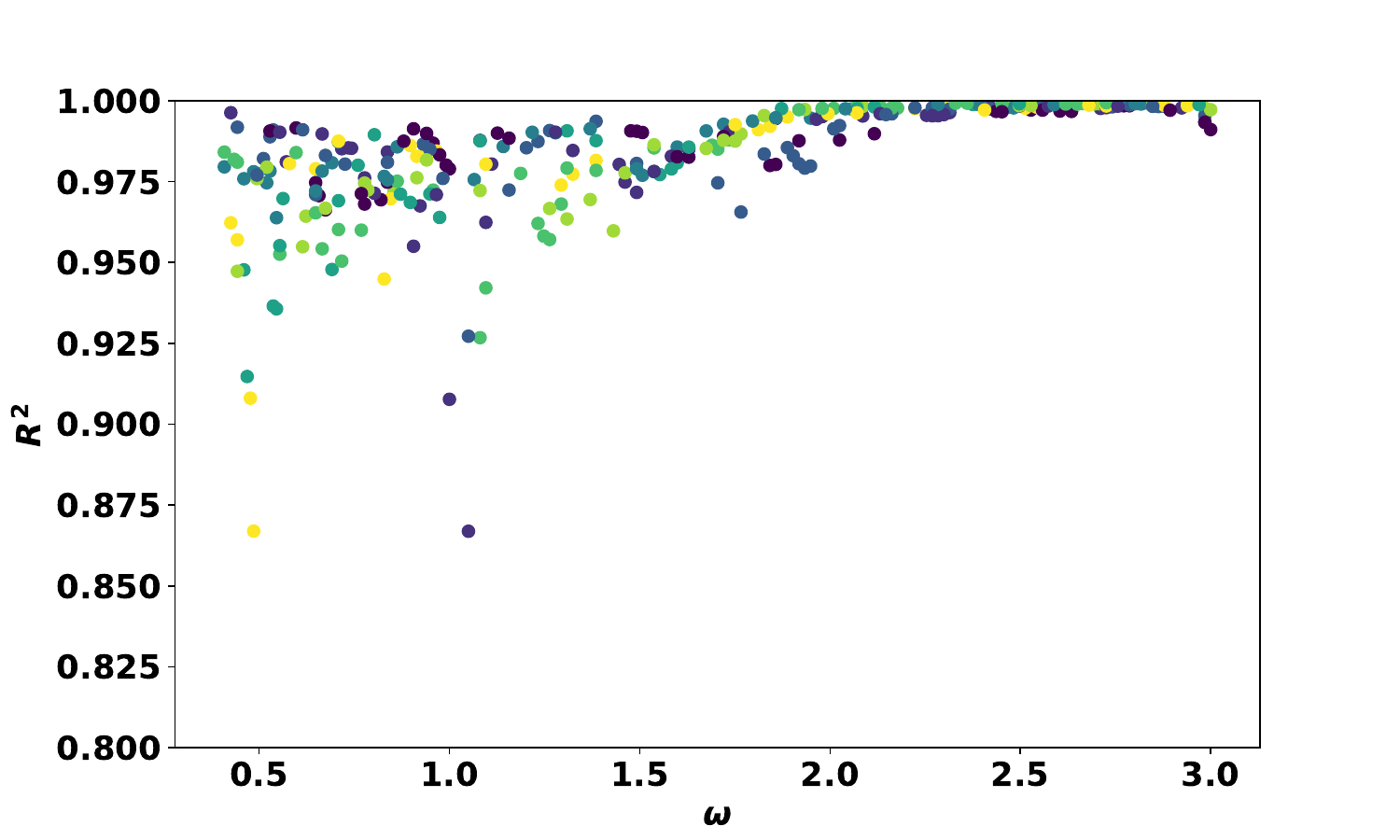}      \caption{\label{fig:Ensemble_multi_amplitudes}%
    Performance illustration of the single KAN model with the intermediate layer width $100$ trained on dataset \ref{case3} in Sec.~\ref{sec:Model}. The figure shows dependence of $R^{2}$ scores on the driving frequency for different amplitudes for the input signals from the test set. Different amplitudes are shown in different colors, with smallest amplitudes corresponding to yellow dots and darker colors for larger amplitudes.    }     
\end{figure}

\begin{figure}[h!]    
\includegraphics[width= \linewidth]{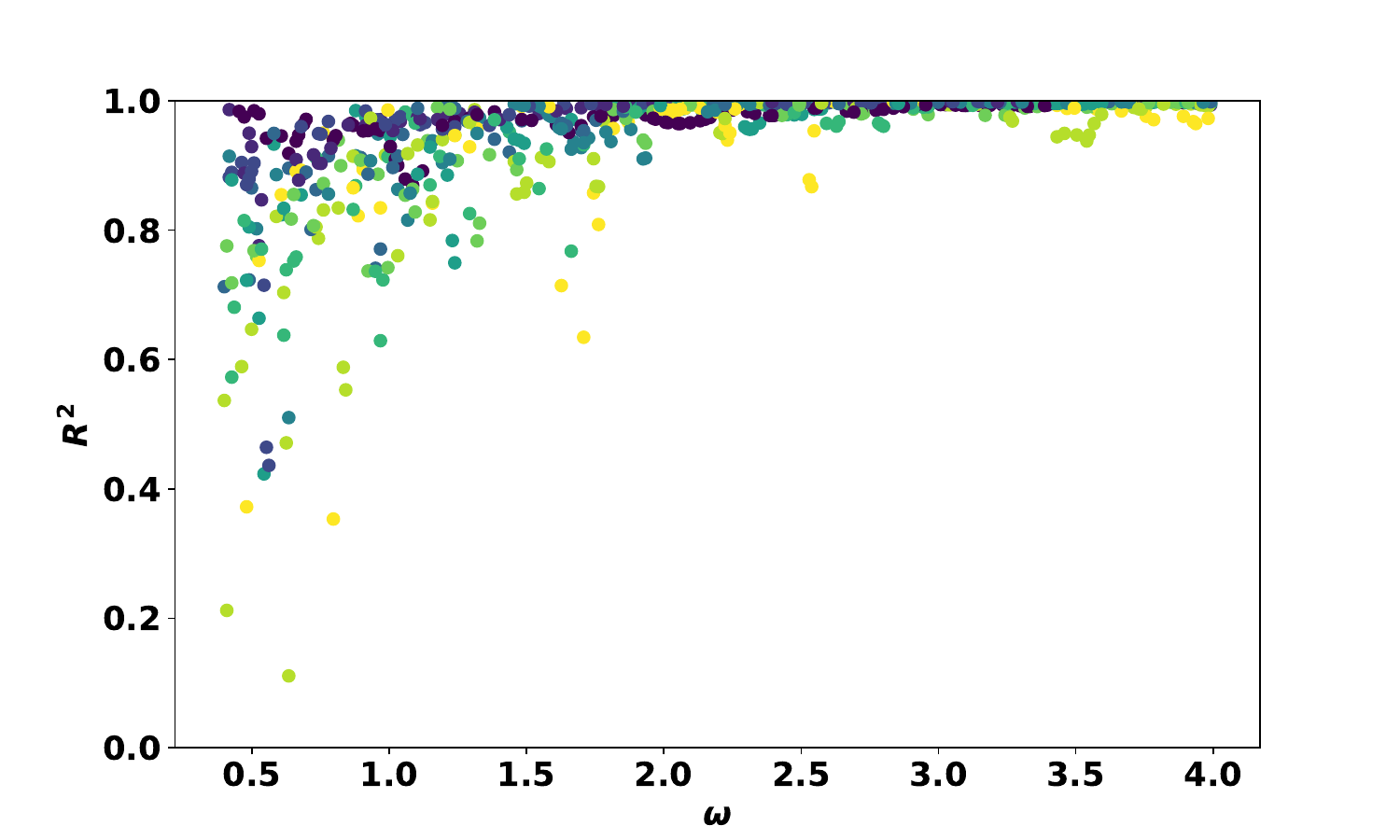}      \caption{\label{fig:Ensemble_multi_amplitudes_LD_A10}%
    Performance illustration for the KAN model with intermediate layer width $1000$ trained on dataset ~\ref{case4} in Sec.~\ref{sec:Model}. See the caption of Fig.~\ref{fig:Ensemble_multi_amplitudes} for further details. Note that one point is excluded from the plot due to $R^{2}<0$. }     
\end{figure}

\begin{figure}[h!]    
\includegraphics[width= \linewidth]{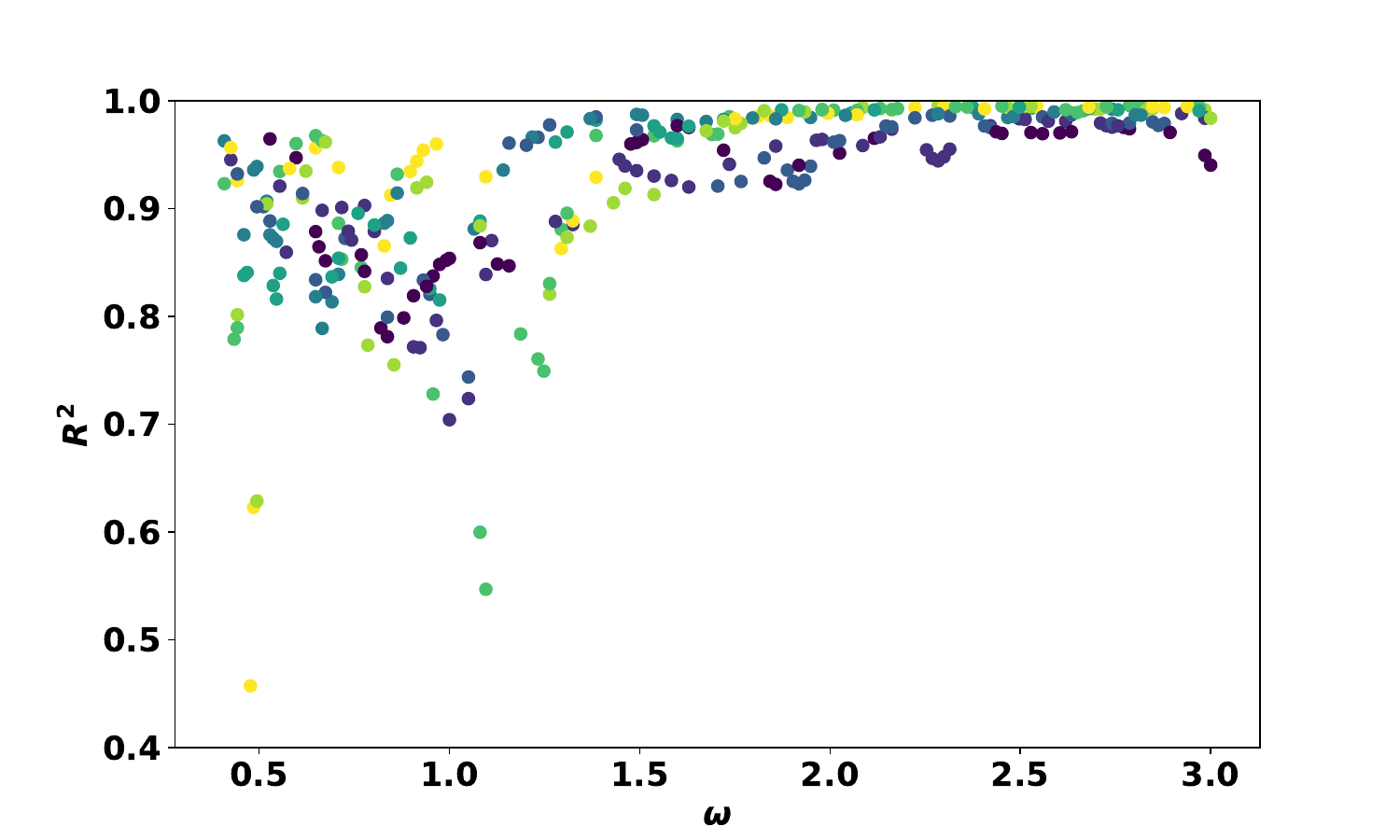}      \caption{\label{fig:Chain_multi_amplitudes}%
    Performance illustration for the chain of KANs model with $a=100$ trained on dataset \ref{case3} in Sec.~\ref{sec:Model}. See the caption of Fig.~\ref{fig:Ensemble_multi_amplitudes} for further details.  }     
\end{figure}

In this subsection, the results for the datasets consisting of multiple amplitudes are presented (datasets \ref{case3} and \ref{case4} in Sec.~\ref{sec:Model}). These datasets contain a much larger number of input signals and the complexity of the output signals is also higher than that for the single-amplitude datasets. 

The results for the single KAN architecture with intermediate layer width 100 for dataset \ref{case3} are shown in Fig.~\ref{fig:Ensemble_multi_amplitudes}. The percentage of test cases with scores above 0.95 is 95\%, while the percentage of scores above 0.9 is 98\%. It can be seen that quite generally the $R^{2}$ scores are very high, at least for sufficiently high  frequencies. The major drawback is that we still need to use quite a large intermediate layer width to achieve good accuracy. 

The results for the most complex dataset \ref{case4} are shown in Fig.~\ref{fig:Ensemble_multi_amplitudes_LD_A10}. Note that here we use $N_\omega=400$ per amplitude, since it was difficult to train the model with only 200 frequencies. The percentage of $R^{2}>0.9$ in the test set is 85\%, while the percentage of $R^{2}>0.95$ is 74\%. The predictions do not demonstrate good accuracy at small frequencies. Besides, there was one point where the prediction was completely unsuccessful, with a negative $R^{2}$ score. Still, taking into account the data complexity, these results can be considered quite promising. 
In Fig.~\ref{fig:Chain_multi_amplitudes} we additionally show the results for dataset \ref{case3} and the chain of KANs architecture.  The percentage of scores with $R^{2}>0.95$ is 52\%, while for $R^{2}>0.9$ it is 70\%.

\subsection{Architecture stability}\label{ArchitectureStability}

An additional question concerns the choice of the optimal KAN architecture. The optimal architecture is the one that has a minimal number of layers as well as minimal widths for the intermediate layers, while still providing sufficiently accurate predictions. Obviously, there are potentially many different combinations of the numbers and widths of layers, and it is difficult to check them all. In this subsection, we focus on the architectures with just one intermediate layer and search for the minimal width for this layer. 

We also note that the results for the training may vary depending on the partitioning of the dataset into the training and test parts. For this reason, we train our models for various different partitions (generally for 10 different partitions) and summarize the results in the form of tables, where we show the percentage of test results with $R^{2}$ scores exceeding the threshold values 0.98, 0.95, and 0.9. 

\begin{table}[h!]
    \centering
\caption{KAN with $a = 10$, $A=2.6$, and $N_\omega=200$.}
\label{tab:A=26,w=10}
    \begin{tabular}{|c|c|c|c|} \hline 
    ~Model No.~&~$R^2>0.98,\,\%$~&~$R^2>0.95,\,\%$~& ~$R^2>0.9,\,\%$~\\ \hline
         1&  67.5&  77.5 & 80.5\\ \hline 
         2& 52.5  &  65 & 77.5 \\ \hline 
         3&  50 &  70& 80\\ \hline 
         4&  57.5 &  70& 77.5 \\ \hline 
         5&  60&  75& 85\\ \hline 
         6&  57.7 &  75& 90\\ \hline 
         7&  62.5 &  75& 87.5\\ \hline 
         8&  80&  82.5& 95\\ \hline 
         9&  70 &  82.5 & 82.5\\ \hline 
         10&  55&  67.5& 77.5\\ \hline
    \end{tabular}
\end{table}

\begin{table}[h!]
    \centering
\caption{KAN with $a = 20$, $A=2.6$, and $N_\omega=200$.}
\label{tab:A=2.6,w=20}
    \begin{tabular}{|c|c|c|c|} \hline 
    ~Model No.~&~$R^2>0.98,\,\%$~&~$R^2>0.95,\,\%$~& ~$R^2>0.9,\,\%$~\\ \hline
         1&  73&  83& 90\\ \hline 
         2&  58&  75& 93\\ \hline 
         3&  53&  75& 90\\ \hline 
         4&  70&  85& 93\\ \hline 
         5&  80&  90& 98\\ \hline 
         6&  58&  83& 90\\ \hline 
         7&  70&  88& 93\\ \hline 
         8&  88&  95& 100\\ \hline 
         9&  65&  85& 90\\ \hline 
         10&  63&  68& 78\\ \hline
    \end{tabular}
\end{table}

\begin{table}[h!]
    \centering
\caption{KAN with $a = 10$, $A=2.6$, and $N_\omega=400$.}
\label{tab:A=2.6,w=10,LD}
    \begin{tabular}{|c|c|c|c|} \hline 
    ~Model No.~&~$R^2>0.98,\,\%$~&~$R^2>0.95,\,\%$~& ~$R^2>0.9,\,\%$~\\ \hline
         1&  50&  76& 91\\ \hline 
         2&  47&  94& 97\\ \hline 
         3&  65&  86& 91\\ \hline 
         4&  65&  89& 94\\ \hline 
         5&  53&  73& 96\\ \hline 
         6&  68&  88& 90\\ \hline 
         7&  41&  74& 94\\ \hline 
         8&  81&  95& 99\\ \hline 
         9&  63&  90& 95\\ \hline 
         10&  56&  80& 91\\ \hline
    \end{tabular}
\end{table}

First, we should note that the minimal stable architecture depends on the dataset. For dataset~\ref{case1}, which has amplitude $A=2.6$, a relatively small intermediate width $a=10$ or $a=20$ is already sufficient to reach good accuracy. From a comparison of Tables~\ref{tab:A=26,w=10} and \ref{tab:A=2.6,w=20} one can see that the accuracy gradually increases with the width of the intermediate layer. Note that the accuracy also depends on the number of frequencies in the dataset. In particular, in Table~\ref{tab:A=2.6,w=10,LD} we show the results with  width $a=10$ trained on dataset \ref{case1} with $N_\omega=400$, rather than  $N_\omega=200$ as in Table~\ref{tab:A=26,w=10}. In this case, width $a=10$ gives the same accuracy level as we obtained previously with $a=20$ for the dataset with $N_\omega=200$  (Table~\ref{tab:A=2.6,w=20}). Still, in both cases, rather small intermediate layer widths were sufficient to reach good accuracy. This is in contrast to dataset~\ref{case2} (with $A=10$), where the accuracy increases very slowly with the intermediate layer width. We show an example of results for dataset \ref{case2} with intermediate width $a=100$ in Table~\ref{tab:A=10,w=100}. Even for such a relatively large width, on average only about $60\%-70\%$ of our results have $R^{2}>0.9$. 

\begin{table}[h!]
    \centering
\caption{KAN with $a = 100$, $A=10$, and $N_\omega=200$.}
\label{tab:A=10,w=100}
    \begin{tabular}{|c|c|c|c|} \hline 
    ~Model No.~&~$R^2>0.98,\,\%$~&~$R^2>0.95,\,\%$~& ~$R^2>0.9,\,\%$~\\ \hline
         1&  58&  63& 75\\ \hline 
         2&  48&  55& 68\\ \hline 
         3&  33&  43& 58\\ \hline 
         4&  45&  53& 75\\ \hline 
         5&  55&  65& 83\\ \hline 
         6&  38&  50& 63\\ \hline 
         7&  38&  50& 63\\ \hline 
         8&  60&  65& 70\\ \hline 
         9&  58&  60& 73\\ \hline 
         10&  48&  50& 70\\ \hline
    \end{tabular}
\end{table}

We can conclude that the optimal architecture for KAN depends rather strongly on the properties of the signal. For complex signals, the network must be sufficiently wide (or deep). 

\emph{Code availability:} All codes used in this study can be found in~\cite{git_kan_ts}. 

\section{Conclusion}

We have developed and validated KAN-Ehrenfest Time-Series Analysis, a physics-informed machine learning framework that combines Kolmogorov–Arnold Networks (KANs) with Ehrenfest's theorems to predict quantum system dynamics while preserving the underlying physics of the model. The resulting KAN-ETS approach addresses a critical limitation in existing time series methods in physics: the tendency to violate physical constraints in favor of empirical accuracy. Our method embeds Ehrenfest’s relations directly into the loss function, additionally regularizing it and resulting in smoother predictions. The code examples are available in Ref.~\cite{git_kan_ts}. 

Our method accurately reconstructs magnetization patterns in Ising chains; over 95\% of test cases achieve an $R^2$ above 0.95 using inputs that sample just 200 equispaced frequencies. KANs  match or outperform traditional models such as Temporal Convolution Networks (TCNs)~\cite{sen_input-output_2025} while requiring 95\% less data and far fewer computational resources. This proves that KANs are both data-efficient and faster to train.

Further, another significant advancement of our framework lies in its treatment of the multi-amplitude problem through prompt-conditioned mapping. Rather than requiring separate models for different drive amplitudes (a computationally expensive approach that limits practical applications), the KAN-ETS framework learns a continuous response surface across all amplitude regimes within a single model, seamlessly interpolating from linear response at small drives to highly nonlinear behavior at extreme amplitudes. This mirrors how large language models handle diverse prompts within a unified architecture, adapting their responses based on input context, rather than requiring separate models for different query types. The approach scales naturally with the available data through curriculum learning that progresses from moderate to extreme amplitudes, creating a robust model that faithfully reproduces material behavior across the entire amplitude spectrum and providing a clear path from controlled laboratory studies to large-scale applications without architectural modifications.


We further enriched our modeling toolkit with the chain of KANs, enforcing strict temporal causality along with the Ehrenfest constraint. Unlike traditional models that predict entire output sequences at once, the chain of KANs decomposes the prediction into a sequence of causally ordered steps. Each output point is predicted using only current and past input values, ensuring that future information does not leak into the prediction, preserving causality. By assigning a separate KAN to each time step, the model allows independent training and flexible deployment of individual components, which is particularly useful for online and real-time applications. Despite this decomposition, the chain of KANs maintains strong predictive performance and offers improved interpretability, making it especially valuable for physics-informed forecasting tasks such as quantum control, spectroscopy, and modeling nonlinear optical systems.

\acknowledgments

This work was supported by Army Research Office (ARO) (grant W911NF-23-1-0288; program manager Dr.~James Joseph). The views and conclusions contained in this document are those of the authors and should not be interpreted as representing the official policies, either expressed or implied, of ARO, or the U.S. Government. The U.S. Government is authorized to reproduce and distribute reprints for Government purposes notwithstanding any copyright notation herein.

\appendix
\label{appendix}
\section{Kolmogorov-Arnold Theorem and Kolmogorov-Arnold Network}

\subsection{Quick Introduction}
The development of DNNs, also referred to as feed-forward neural networks or multilayer perceptrons (MLPs), and Convolutional Neural Networks (CNNs) has played a foundational role in the evolution of modern machine learning architectures. While DNNs serve as a general-purpose framework applicable to a wide variety of learning tasks, CNNs are architecturally specialized for data with spatial structure, particularly images.

In traditional MLPs and CNNs, activation functions are essential nonlinear components applied after each linear transformation or convolution. Their role is to introduce nonlinearity into the model, allowing the network to approximate complex, non-linear mappings rather than collapsing into a series of linear operations. Common activation functions such as ReLU, Tanh, or GELU are fixed and applied uniformly across the network. Without them, even a deep stack of layers would behave like a single linear transformation. In contrast, KANs depart from this paradigm by embedding nonlinearity directly in the connections between neurons. Instead of static scalar weights followed by a global activation, each connection in a KAN is a learnable univariate function, often implemented as a B-spline that maps input values in a data-driven, nonlinear fashion (see Fig.~\ref{figkan}).

\subsection{Kolmogorov–Arnold Representation Theorem in KANs}

KAN is a neural architecture that leverages the functional decomposition guaranteed by the Kolmogorov–Arnold representation theorem \cite{arnold_representation_2009}. The theorem is stated below:

Let \(n\in\mathbb{N}\) and \(I=[0,1]\). For any continuous function
$$
f \in C\bigl(I^n,\mathbb{R}\bigr),
$$
there exist continuous “inner” functions
$$
\phi_{i,j}\colon I \to \mathbb{R},
\quad
i=0,1,\dots,2n,\;j=1,2,\dots,n,
$$
and continuous “outer” functions
$$
\psi_i\colon \mathbb{R} \to \mathbb{R},
\quad
i=0,1,\dots,2n,
$$
such that for every \(\mathbf{x}=(x_1,\dots,x_n)\in I^n\) one has
\begin{equation}\label{eq:KAN}
f(x_1,\dots,x_n)
\;=\;
\sum_{i=0}^{2n}
\psi_i\!\Bigl(\sum_{j=1}^n \phi_{i,j}(x_j)\Bigr).
\end{equation}

In 1900, David Hilbert famously asked whether any continuous function of three or more variables could be built from just continuous functions of two variables \cite{morris_hilbert_2020}. Then, in a stunning breakthrough in 1957, Andrey Kolmogorov proved that any continuous function of several variables can be exactly expressed as a finite sum of single-variable functions composed with addition, and soon thereafter Vladimir Arnold refined and streamlined that construction \cite{arnold_representation_2009}. This surprise solution not only laid Hilbert's question to rest, but also inspired today’s Kolmogorov-Arnold Networks, which use trainable univariate subnetworks to capture complex, high-dimensional behaviors.

\subsection{KANs as Networks}\label{sec:KAN}
In this section, we will present the underlying mathematics behind KAN and how it differs from traditional DNN. We will first present how each neuron functions in DNN and then move on to how, in a KAN, each connection itself implements a learnable univariate mapping.
In traditional DNN, the input data is initially received by the input neurons, which then undergoes a pre-activation transformation where the inputs are linearly combined with learned weights and adjusted by bias parameters. Subsequently, this weighted sum passes through a neuron and is subjected to a non-linear activation function $f$. Formally, let $\{x_i\}_{i=1}^n$ denote the inputs to a given neuron and $\{w_i\}_{i=1}^n$, $b$ the corresponding weights and bias. The neuron first computes the \emph{pre‐activation}
\begin{equation}
     z \;=\; \sum_{i=1}^{n} w_i\,x_i \;+\; b \,.
\end{equation}
To introduce nonlinearity, which is essential for the network’s ability to approximate complex functions, the pre‐activation value is then passed through an element‐wise activation function $f$, yielding the \emph{post‐activation}
\begin{equation}
     a \;=\; f(z).
\end{equation}


\begin{figure}[ht]
  \centering
  \begin{subfigure}{\Large(a)}
    \centering
    \includegraphics[width=\linewidth]{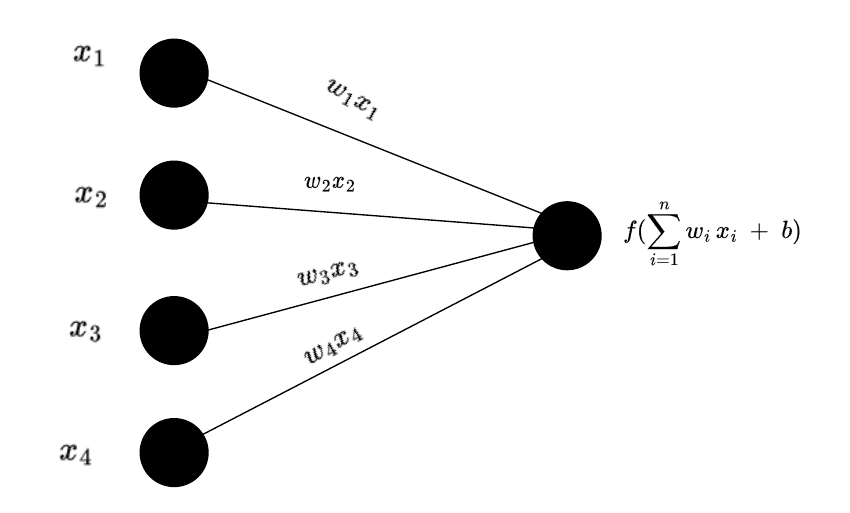}
    \label{fig:kan-dnn}
  \end{subfigure}

  \begin{subfigure}{\Large(b)}
    \centering
    \includegraphics[width=\linewidth]{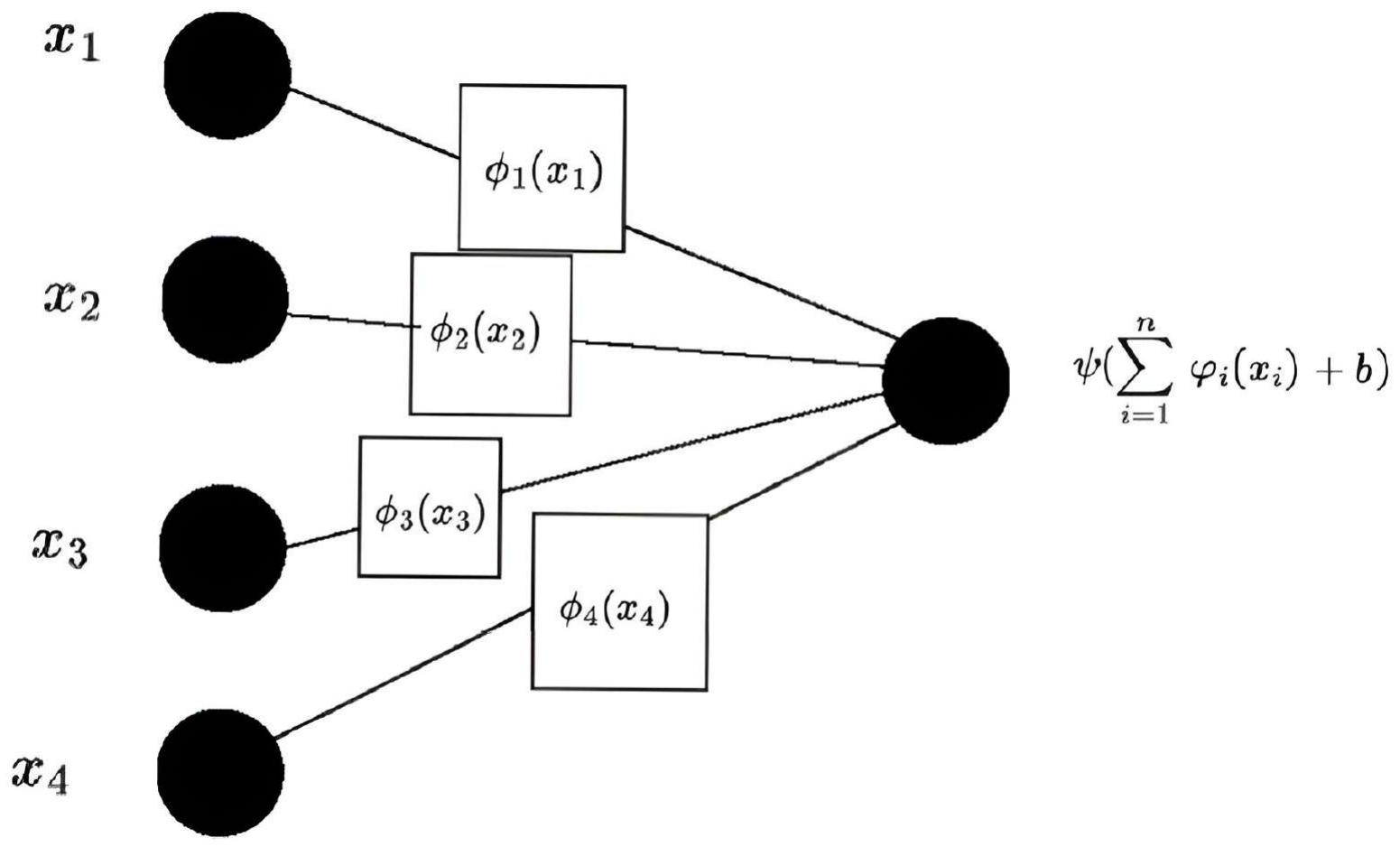}
    \label{fig:kan-kan}
  \end{subfigure}

  \caption{Comparison of (a) a traditional deep neural network and (b) a Kolmogorov–Arnold Network.}
  \label{fig:kan}
\end{figure}

In KAN, each input connection applies a learnable univariate function $\varphi_i:\mathbb{R}\to\mathbb{R}$, replacing scalar weight parameters and embedding non-linearity at every edge. Formally, let $\{x_i\}_{i=1}^n$ denote the inputs to a neuron, $\{\varphi_i\}_{i=1}^n$ the corresponding edge functions, and $b$ the bias term. The neuron then computes the \emph{pre-activation}
\begin{equation}
  z \;=\; \sum_{i=1}^n \varphi_i(x_i)\;+\;b,
\end{equation}
without requiring a separate activation step, since each $\varphi_i$ is itself nonlinear by design. Optionally, to fully mirror the classical superposition form, one may apply a \emph{outer} univariate function \(\psi:\mathbb{R}\to\mathbb{R}\), which results in
\begin{equation}
  a = \psi(z).
\end{equation}

By composing $L$ such layers, denoting the \(\ell\)th layer’s function matrix by \(\phi^{(\ell)}\), one immediately recovers the general deep KAN mapping  
\begin{equation}
     \mathrm{KAN}(x)
  =\bigl(\phi^{(L-1)}\circ\phi^{(L-2)}\circ\cdots\circ\phi^{(0)}\bigr)(x),
\end{equation}
which naturally extends the single‐neuron formulation to architectures of arbitrary depth and width. 

We conclude this section with a brief overview of the different types of KAN that have been proposed. The KAN architecture has rapidly evolved into several specialized variants such as FastKAN, Efficient-KAN, Wavelet-based KAN (Wav-KAN), T-KAN, etc., each tailored to address specific computational challenges or to suit distinct application domains~\cite{somvanshi_survey_2025}. In this article, we tend to use Efficient-KAN~\cite{Blealtan2024} and Wav-KAN \cite{WAV-KAN}. Efficient‑KAN optimizes the original KAN architecture for memory and computational efficiency, making it more practicable for large-scale or resource‑constrained settings. On the other hand, Wav‑KAN enhances the representational power and interpretability by integrating multi-resolution wavelet activation functions, well suited for data with rich frequency structure.

\subsection{KAN Architecture Notation}\label{subsec:KAN_arch}

In this article, we represent a KAN model by the four‐tuple
$$
[I,\,a,\,b,\,O],
$$
(the same notation being used in our implementation code) Here, $I$ denotes the dimensionality of the input layer (i.e., the number of input features), $a$ and $b$ denote the widths of the first and second hidden layers respectively, and $O$ denotes the dimensionality of the output layer (i.e., the number of output features).

For the specific configuration
$$
[500,\,100,\,100,\,500],
$$
we have $I=500$, meaning each data example is a univariate time series of length 500 (treated as a 500‐dimensional input vector). The hidden layers each contain 100 nodes ($a=b=100$), providing a compact nonlinear embedding, and $O=500$, so the network reconstructs or forecasts a 500‐step output sequence.

\section{Failure Modes of Temporal Convolution Networks}
\label{TCN_failure}
In Ref.~\cite{sen_input-output_2025}, the input → output mapping for both the transverse and non‐integrable Ising models is formulated as follows.  We drive a quantum spin chain of length \(N\) with a monochromatic magnetic field of fixed amplitude $A$,
\begin{equation}
    h^{(i)}(t) = A \sin\bigl(\omega^{(i)}\,t\bigr)\, \quad
 \text{with} \quad i=1,\dots,n,
\end{equation}
where \(n=3700\).  The corresponding optical output is the total magnetization  
\begin{equation}
    Y^{(i)}(t) \;=\; \sum_{k=1}^N \langle\sigma_x^{(k)}(t)\rangle\,.
\end{equation}

Then, discretizing time into \(T=512 \) uniform steps of size \(\delta t\) transforms each continuous trajectory into a pair of \(T\)-dimensional vectors  
\begin{equation}
\begin{split}
\text{Input: }h^{(i)}_k &= A \sin\bigl(\omega^{(i)}\,k\,\delta t\bigr),\\
\text{Output: }Y^{(i)}_k &= Y^{(i)}\bigl(k\,\delta t\bigr),
\end{split}
\end{equation}
where $k = 1,2,\dots,T$. Thus, the full dataset can be written as  
\begin{equation}
    \bigl\{\,\mathbf{h}^{(i)} \to \mathbf{Y}^{(i)}\bigr\}_{i=1}^n,\quad
\mathbf{h}^{(i)},\mathbf{Y}^{(i)}\in\mathbb{R}^{T},
\end{equation}

\begin{figure}[ht]
    \centering
    \includegraphics[width=1\linewidth]{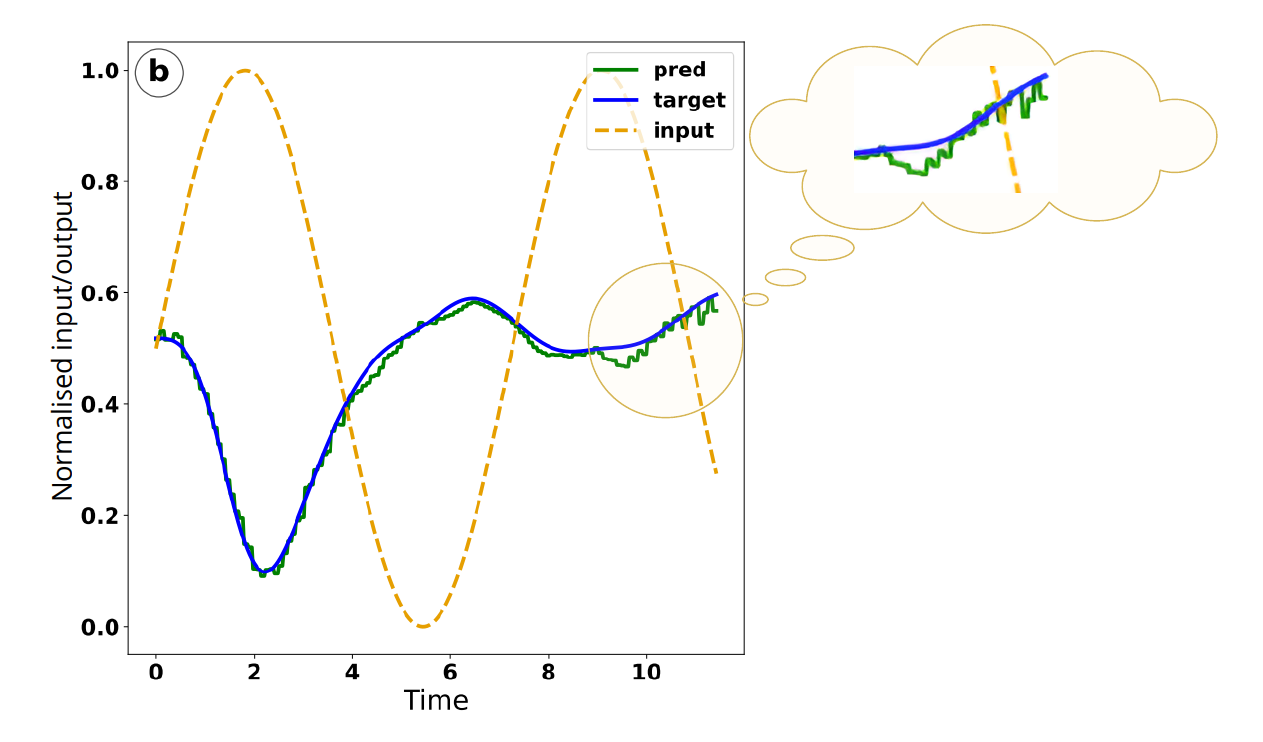}
    \caption{Predicted output signal using a TCN model. The TCN output (green) compared with the target signal (blue), for a given input (orange), exhibits step-size residual oscillations, highlighted in the inset, which remain unresolved even after training on large data samples.}
    \label{TCNLIMITATION}
\end{figure}

It was observed that when the number of input frequencies falls below this threshold, the TCNs, regardless of architectural depth or filter size, tend to become trapped in poor local minima during training, resulting in plateau loss and consequently inadequate input–output fitting. Moreover, owing to the inherently large receptive fields and repeated convolutional operations required to process such high-dimensional data, the training time for TCNs scales unfavorably. In contrast, we will see that KANs achieve comparable or superior fit performance with as few as 200 frequency samples, while incurring substantially lower computational overhead. In particular, where a TCN baseline required approximately 3,700 data samples to reach a given level of precision, KANs reached that regime with only 200, underscoring both their data efficiency and reduced training cost.
Another unresolved issue in~\cite{sen_input-output_2025} lies with tiny step-size residual oscillations in the predicted output signal (see Fig.~\ref{TCNLIMITATION}). Although the original article does not investigate the source of these oscillations/fluctuations, in the present work we demonstrate that they vanish when employing KANs.

\bibliography{references}

\end{document}